\documentclass[journal]{new-aiaa}
\usepackage{caption, booktabs}
\usepackage{subcaption}
\usepackage{graphicx} 
\usepackage{longtable,tabularx}
\usepackage{xcolor} 
\usepackage{colortbl}
\usepackage{color}

\newcommand{\E}{\mathbb{E}}

\newcommand{\cP}{\mathcal{P}}
\newcommand{\cA}{\mathcal{A}}

\DeclareMathOperator*{\argmax}{arg\,max}

\newtheorem{helptheorem}{Theorem}{} 
 
\newtheorem{helplemma}[helptheorem]{Lemma} 

\newtheorem{helpcorollary}[helptheorem]{Corollary} 
 
\newtheorem{helpexample}[helptheorem]{Example} 
 
\newtheorem{helpproposition}[helptheorem]{Proposition} 
 
\newtheorem{helpremark}[helptheorem]{Remark} 
 
\newtheorem{helpdefinition}[helptheorem]{Definition}
 
\newtheorem{helpassumption}[helptheorem]{Assumption}

\newtheorem{helpproblem}[helptheorem]{Problem}

\newenvironment{problem} 
{\vskip.1cm\begin{helpproblem}} 
{\end{helpproblem}\vskip.1cm}

\title{Assessing Autonomous Inspection Regimes: Active Versus Passive Satellite Inspection\footnote{The views expressed are those of the author and do not necessarily reflect the official policy or position of the Department of the Air Force, the Department of Defense, or the U.S. government. Approved for public release; distribution is unlimited. Public Affairs approval \#AFRL-2024-6671.}}
 
\author{Joshua Aurand\footnote{joshua.aurand@verusresearch.net}, Christopher Pang, Sina Mokhtar, Henry Lei, and Steven Cutlip}
\affil{Verus Research}
\author{Sean Phillips}
\affil{Air Force Research Laboratory, Kirtland AFB, NM. 87117}

\begin{document}
 
\maketitle

\begin{abstract}

This paper addresses the problem of satellite inspection, where one or more satellites (inspectors) are tasked with imaging or inspecting a Resident Space Object (RSO) due to potential malfunctions or anomalies. Inspection strategies are often reduced to a discretized action space with predefined waypoints, facilitating tractability in both classical optimization and machine learning-based approaches. However, this discretization can lead to suboptimal guidance in certain scenarios. This study presents a comparative simulation to explore the tradeoffs of passive versus active strategies in multi-agent missions. Key factors considered include RSO dynamic mode, state uncertainty, unmodeled entrance criteria, and inspector motion types. The evaluation is conducted with a focus on fuel utilization and surface coverage. Building on a Monte-Carlo based evaluator of passive strategies and a reinforcement learning framework for training active inspection policies, this study investigates conditions under which passive strategies, such as Natural Motion Circumnavigation (NMC), may perform comparably to active strategies like Reinforcement Learning based waypoint transfers.
\end{abstract}

\section*{Nomenclature}

{\renewcommand\arraystretch{1.0}
\noindent\begin{longtable*}{@{}l @{\quad=\quad} l@{}}
$\mathcal{A}$ & Set of actions for inspection strategy\\
$\mathcal{F}$ & Reference frame with origin $\mathcal{O}$\\
$f_{\text{Hill}}$ & Analytical solution for agent trajectories in the Hill's frame\\
$I$   & Inertia matrix, $I \in \mathbb{R}^{3\times3}$ - under the no products of inertia assumption $I \in \mathbb{R}^{3} = [I_{xx}, I_{yy}, I_{zz}]$ \\
$\mathcal{I}_d$ & Set of inspection agents\\
$I_{\cP}$ & Inspection state \\
$IG(\bar{v})$ & Cumulative information gain\\
$M$ & Information threshold, $M\in[0,1]$ \\
$\overline{m}$ & Cardinality of set of allowable viewpoints. \\
$m_i$ & Mass of the i-th inspector satellite \\
$n$ & Number of inspector agents. \\
$O$ & Observation probability function \\
$\mathcal{O}$ & Agent observations \\
$\mathcal P$ & Points of interest on inspection target \\
$\textbf{p}(v)$ & Visible points of the target point cloud within an image from viewpoint $v$\\
$Q_{i}^{\pi_i}$ & Action value function \\
$q$ & Attitude represented as a quaternion, $q \in SO(3)$ (the 3D rotation group) \\
$r_{i}$ & Position of agent $i$ in ECI\\
$\delta r_{i}$ & Relative position of agent $i$ to RSO in Hill Frame\\
$r_0$ & Orbital radius of inspection target orbit \\
$\mathcal{S}$ & Dynamics system state\\
T & Mission time horizon\\
$\Delta T(a_i, a_j)$ & Expected time-of-flight (TOF) for a single burn trajectory between points $a_i$ and $a_j$\\
$u_i (t)$ & Control output as thrust \\
$\Delta V(a_i, a_j)$ & Fuel usage of a transfer between the points $a_i$ and $a_j$ measured in (m/s)\\ 
$\Delta v$  & Change in velocity, comes from integrating the CWH equations and is used as a surrogate for fuel cost \\
$\mu$ & Standard gravitational parameter of Earth \\
$\pi_{i}^{hl}$ & High level policy \\
$\pi_{i}^{ll}$ & Lower level policy \\
$\sigma_{\mu}$ & Spatial gravity parameter \\
$\tau$ & First hitting time of inspection completion \\
$\omega$ & Vector of angular velocity components for RSO, $\omega \in \mathbb{R}^{3} = [\omega_x, \ \omega_y, \ \omega_z]^T$ \\
$\otimes$ & Quaternion multiplication \\

\end{longtable*}}

\section{Introduction}
This paper considers the problem of satellite inspection which includes the broad scenario where one or more satellites (inspectors) aim to image (or inspect) some portion of a Resident Space Object (RSO) due to potential malfunctioning or mishap anomalies. The inspection task faces several challenges, including the need to plan rapidly in a nonlinear,  highly constrained, and uncertain environment. Using a team of satellites to perform the inspection has the potential to increase task efficiency with many existing solutions actively exploiting this; see \cite{Lei22,Bernhard20,Nakka21,Phillips2022, JSR24}. In such scenarios, it is often the case where complexity of planning is directly proportional to the strength of modeling assumptions used for environmental simulation. This provides a tacit tradeoff between tractability and realism. In \cite{Bernhard20} and \cite{Nakka21} solutions are restricted to parking orbits along pre-computed elliptical natural motion trajectories, which can increase both the time to complete an inspection and the number of inspecting agents required to get full sensor coverage of the target. In \cite{Phillips2022}, simplified Clohessy-Wiltshire equations and a Lyapunov-based closed-loop control method were used to converge a multi-satellite spacecraft system to a formation around a chief agent. In \cite{oestreich21, Dor18, hays2023}, the focus is on relative state and pose estimation from realistic images. The inspection task is then formulated as a simultaneous localization and mapping (SLAM) problem; \cite{GrangePLANS2023} then investigates the consensus of pose for multiple agents. Neither imposes major assumptions on the RSO itself, but both assume a fixed trajectory for the inspecting agent. The stage of trajectory optimization is not explicitly considered. Meanwhile, in works that consider trajectory design itself, the problem of relative state estimation is often simplified by assuming the target is cooperative with only a single inspecting agent; for example, in \cite{woffinden07}. In \cite{Maestrini22}, trajectory optimization is combined with an unknown, non-cooperative target but again using only a single inspecting agent. Alternatively, \cite{miller2023SciTech} considers a multi-satellite autonomous rendezvous proximity operation and docking (ARPOD) problem posed as a game with lexicographic cost functions. This solution requires onboard trajectory sampling and simulation that quickly becomes computationally expensive with a large potential for excessive power draw. 

Complementary to these approaches, Monte Carlo analysis has become a valuable tool for simulation and planning. This includes problems such as satellite conjunction analysis \cite{Vries2010}, satellite stabilization \cite{Omar2021}, and satellite docking \cite{Stesina2021}. The dynamic behavior of satellites during the active `intercept and dock' problem has been considered in both cooperative \cite{Stesina2021} and tumbling circumstances \cite{Albee2021} \cite{Zampato2013} \cite{Lampariello2021}. With regards to trajectory design for inspection of a tumbling RSO, this has been effectively used in both single agent\cite{Setterfield2017} and multi-agent scenarios \cite{Nakka2021} \cite{Bernhard2020}. In both active (strategies may be updated during the course of the mission) and passive (strategy is fixed at mission initialization) inspection scenarios, being able to predict the dynamic behavior of the target is important because rigid body dynamics can be chaotic and small changes in initial conditions can lead to very different rotational dynamics; see \cite{Elipe1997} \cite{Richter2006}. Even given an accurate model, predictions are usually uncertain because initial conditions are often taken from noisy state measurements. One means of accounting for uncertainty in initial conditions is to perform a Monte Carlo analysis within a prespecified region defined by an estimate of initial condition. This can then be used to help hedge against uncertainty in trajectory design efficacy by providing a relatively simple means of situational analysis.

With a range of scenarios and configurations being studied, it is prudent to investigate which modeling assumptions are most impactful and under which circumstances. This will help the community better understand trade-offs between simplifying assumptions and strategy performance. The exposition that follows will focus specifically on assessing the impact of imposing a ``passivity" constraint on translational guidance generation. We provide a comparative simulation study to better understand and define regimes in which a multiagent inspection mission is more efficiently done in a passive or active manner. Key parameters of interest include the efficiency trade-offs between environment assumptions, state uncertainty, number of inspectors, and inspector motion type on measures that include fuel usage, inspection time, and inspection performance. This work leverages and extends on our previous two papers: 1) A Monte Carlo-based evaluator of passive multi-agent inspection strategies \cite{SC2024} and 2) a reinforcement learning-based framework for training multiagent inspection policies \cite{JSR24}. The extension of our previous work presented in this paper uses \cite{SC2024} as a comparison point for \cite{JSR24}. Based on different assumption primitives (specified in Table~\ref{tab:ValidationFactors}, we assess situations in which a passive strategy's (such as a Natural Motion Circumnavigation: NMC) performance is comparable to that of an active strategy (such as a Reinforcement Learning policy trained to perform waypoint transfers).

\section{Background and Problem Statement}

The autonomous inspection task considers the inspection of an on-orbit RSO conducted via one or multiple inspecting satellites equipped with sensors performing surface area scans within a certain range of the target. Inspection completion can be measured through a visibility calculation on predefined points of interest (POIs) on the surface of the target. The global environment in which the inspection task is carried out consists of the following key components: frame specification, member dynamics, and deputy guidance strategy. This is summarized below with an accompanying definition of the formal problem statement.

\subsection{Environment Specification}
\subsubsection{Translational Dynamics}
Following the notation in \cite{Phillips_ST24}, let $\mathcal{F}_{\rm E}$ be the \textit{Earth Centered Inertial (ECI) frame} which is defined by the orthogonal unit vectors $\hat{\imath}_{\rm E}, \hat{\jmath}_{\rm E}, \hat{k}_{\rm E}$ whose origin is denoted $\mathcal{O}_{\rm E}$. The $\hat{\imath}_{\rm E}$ direction is pointed along the vernal equinox, the $\hat{k}_{\rm E}$ direction is pointed along the celestial north pole of Earth, and the $\hat{\jmath}_{\rm E}$ direction completes the right-handed-reference frame. Consider $n\in\mathbb{N}$ with $n>0$ deputy spacecraft orbiting Earth and define the deputy index set $\mathcal{I}_{\rm d} \triangleq \{1, \dots, n \}$. The RSO is denoted by index $i=0$. Thus, the full agent index set is defined $I \triangleq I_{\rm d} \bigcup \{ 0 \}$. Then, for $i \in \mathcal{I}$, let $\mathcal{F}_i$ be the $i$th spacecraft's \textit{body-fixed frame} defined by the orthogonal unit vectors $\hat{\imath}_{i}, \hat{\jmath}_{i}, \hat{k}_{i}$ whose origin is denoted $\mathcal{O}_{i}$.

Let $\mathcal{F}_{\rm H}$ denote \textit{Hill's frame} defined by the orthogonal unit vectors $\hat{\imath}_{\rm H}, \hat{\jmath}_{\rm H}, \hat{k}_{\rm H}$ whose origin is defined such that $\mathcal{O}_{\rm H} = \mathcal{O}_0$. Thereby, the $\hat{\imath}_{\rm H}$ direction is pointed along the vector $\mathcal{O}_0$ relative to $\mathcal{O}_{\rm E}$ (i.e., the radial zenith of the RSO with respect to the ECI frame). The $\hat{k}_{\rm H}$ direction is pointed along the angular momentum vector of the RSO, and the $\hat{\jmath}_{\rm H}$ direction completes the right-handed-reference frame.

For $i \in \mathcal{I}$, let $\mathcal{O}_i$ relative to $\mathcal{O}_{\rm E}$  be the the $i$th agent's position relative to the ECI frame which is denoted by the vector $\vec{r}_i$. Resolve $\vec{r}_i$ in $\mathcal{F}_{\rm E}$ and write in matrix form as $r_i \in \mathbb{R}^3$. The dynamics of the $i$th spacecraft resolved in $\mathcal{F}_{\rm E}$ can be written as 
\begin{align}
\ddot{r}_i (t) = -\frac{\mu}{|| r_i(t) ||^3} r_i(t) + \frac{1}{m_i} \left( f_{{\rm p},i}(t) +  f_{{\rm e},i} (t) \right), \label{two:body}
\end{align}
where for $t \geq 0$, $\ddot{r}_i(t) \in \mathbb{R}^3$ is the second time-derivative with respect to $\mathcal{F}_{\rm E}$ of $r_i$; $\mu \in \mathbb{R}$ is the Earth's gravitational parameter; $m_i \in \mathbb{R}$ is the spacecraft's mass; $f_{{\rm p},i}(t) \in \mathbb{R}^3$ are the environmental perturbation forces acting on the spacecraft; $f_{{\rm e},i}(t) \in \mathbb{R}^3$ is an external force acting on the spacecraft resolved in $\mathcal{F}_{\rm E}$; and $r_i(0), \dot{r}_i(0), f_{{\rm p},i}(0), f_{{\rm e},i}(0)$ are the initial conditions. Relative motion of the $i$th deputy relative to the RSO is defined as $\delta \vec{r}_i \triangleq \vec{r}_i - \vec{r}_0$. We resolve $\delta \vec{r}_i$ in $\mathcal{F}_{\rm H}$ and notate in matrix form as $\delta r_i \in \mathbb{R}^3$. To express the relative motion dynamics, consider the following assumptions:
\begin{enumerate}[itemindent=5pt,label=(A{\arabic*})]
    \item \label{A1}
    The RSO is translationally unactuated.

    \item \label{A2}
    The RSO is in a circular orbit.

    \item \label{A3}
    For all $i \in I$, $\delta r_i \in [0, r_{\rm max}]^{3}$, where $r_{\rm max} \in \mathbb{R}_{+}$ is a fixed constant. 
\end{enumerate}
It follows from \ref{A1}--\ref{A3}, that the relative motion dynamics of the $i$th deputy resolved in Hill's frame can be linearized with an error proportional to $r_{\rm max}$ via:
\begin{align}
    \begin{bmatrix}
        \delta \dot{r}_i (t) \\
        \delta \ddot{r}_i (t)
    \end{bmatrix} = 
    \begin{bmatrix}
        0 & 0 & 0 & 1 & 0 & 0 \\
        0 & 0 & 0 & 0 & 1 & 0 \\
        0 & 0 & 0 & 0 & 0 & 1 \\
        3 \sigma_{\mu}^2 & 0 & 0 & 0 & 2 \sigma_\mu & 0 \\
        0 & 0 & 0 & -2 \sigma_{\mu} & 0 & 0 \\
        0 & 0 & -\sigma_{\mu}^2 & 0 & 0 & 0 
    \end{bmatrix}
    \begin{bmatrix}
        \delta r_i (t) \\
        \delta \dot{r}_i (t)
    \end{bmatrix} + 
    \begin{bmatrix}
        0 & 0 & 0 \\
        0 & 0 & 0 \\
        0 & 0 & 0 \\
        \frac{1}{m_{i}} & 0 & 0 \\
        0 & \frac{1}{m_{i}} & 0 \\
        0 & 0 & \frac{1}{m_{i}}
    \end{bmatrix}
    u_{i} (t), \label{eqn: hcw}
\end{align}
where for $t \geq 0$,  $\delta \dot{r}_i(t) \in \mathbb{R}^3$ and $\delta \ddot{r}_i(t) \in \mathbb{R}^3$ are the first and second time-derivatives with respect to $\mathcal{F}_{\rm H}$ of $\delta r_i$; the spatial gravity parameter $\sigma_\mu = \sqrt{\mu / || r_0 ||^3} \in \mathbb{R}$; $u_{i}(t) \in \mathbb{R}^3$ is the desired force input resolved in $\mathcal{F}_{\rm H}$; and $\delta r_i(0), \delta \dot{r}_i(0), u_{i}(0)$ are the initial conditions. For the purposes of notation and for a fixed thrust policy $u_{i}(t)$, we let the forward-integrated linearized dynamics be notated through
\begin{equation}\label{eqn: Hill position}
    \delta r_{i}(t) = \delta r_{i}(0) + \int_{0}^{t}\delta \dot{r}_i (s)ds = f_{\text{Hill}}(t, u_{i}(t))
\end{equation}
where $\delta \dot{r}_i (s)$ satisfies eq.~\eqref{eqn: hcw} under $u(t)$. It should be noted that $\sigma_\mu$ is constant due to assumption~\ref{A1} and \ref{A2}.

\subsubsection{Illumination}
We follow the basic model of \cite{vanWijk23} with added support for non-spherical surfaces through the hidden point removal algorithm in \cite{Katz07}. Primary assumptions for illumination include:
\begin{enumerate}[itemindent=5pt,label=(A{\arabic*})]\addtocounter{enumi}{3}
    \item \label{A4}
    The sun is the only illumination source. The Earth is the only illumination obstruction.

    \item \label{A5}
    The sun is considered to be positionally static in $\mathcal{F}_{\rm E}$ and remains in the orbital plane of the RSO during inspection.

    \item \label{A6}
    The RSO surface is homogenously defined with respect to ambience, diffusivity, specularity, and shininess properties.
\end{enumerate}

\subsection{Guidance Categories and Action Specification}
For the purposes of this paper, there are two distinct hierarchies of translational control for each spacecraft: High-Level (HL) guidance and Low-Level (LL) motion control. HL guidance specifies translational goals for the spacecraft to move towards while the LL motion controller is responsible for calculating a thrust control policy $u_{i}(t)$ that will actually update relative vehicle position according to \eqref{eqn: Hill position}. We fix a discrete set $\cA := \{a_{1},\dots,a_{m}\}$ of actions for the deputy spacecraft HL guidance planner. This represents decision primitives used to analyze implications of different forms of trajectory design on the solution of related inspection problems. Herein, there are three action set specifications considered for comparison: \ref{subsec: waypoint sequence}, \ref{subsec: waypoint hold}, and \ref{subsec: NMC Hold}. Each of these is designed 
around a common control rule for specifying trajectory generation; originally specified in \cite{JSR24}. In particular, for two arbitrary waypoints $(a,\hat{a})\in\mathcal{F}_{\rm H}\times\mathcal{F}_{\rm H}$ the fuel cost and transfer time associated with movement between $a\mapsto \hat{a}$ is calculated according to a basic heuristic allowing for deterministic transfer costs consistent with low-level single-burn thrust control. We assume agents move at the approximate rate needed to maintain a \emph{closed} natural motion trajectory (NMT) connecting viewpoints; determined by the internal angle between $(a,\hat{a})$ specified in $\mathcal{F}_{\rm H}$. A natural motion circumnavigation (NMC), also called a {\it parking orbit}, is a zero-fuel stable orbit defined from state initialization of $\delta r_{i}(0) = (x_{i}(0),y_{i}(0), z_{i}(0))$ satisfying
\begin{equation}\label{eqn: NMC_init}
    y_{i}(0) = 0, \dot{x}_{i}(0) = 0, \dot{y}_{i}(0)=-2\sqrt{\mu / R^3}x_0, \text{ and } \dot{z}_{i}(0) = 0.
\end{equation}
The natural time-of-flight (TOF) for an NMC is simply one orbital period. This motivates a time heuristic for waypoint transfer defined below in \eqref{eqn:LL_proxy}. Rather than requiring an orbital period between information captures in waypoint parking actions, we instead implement a rule defined according to half the time needed to traverse to the nearest neighboring waypoint. More specifically, the TOF is determined by:
\begin{equation}\label{eqn:LL_proxy}
	\Delta T(a,\hat{a}) := 
	\begin{cases}
		\arccos\left(\frac{a\cdot \hat{a}}{\|a\| \|\hat{a}\|}\right)/2\pi\sigma_{\mu}, & a\neq \hat{a}\\
	\min\limits_{\mathbf{x},\mathbf{y}\in\cA}\arccos\left(\frac{\mathbf{x}\cdot \mathbf{y}}{\|\mathbf{x}\| \|\mathbf{y}\|}\right)/4\pi\sigma_{\mu}, &  a=\hat{a}
	\end{cases}.
\end{equation}

The velocity needed to successfully transfer from  $a$ to $\hat{a}$ along an NMT with TOF $\Delta T(a,\hat{a})$ is given by $\mathbf{v}_{0}\left(a,\hat{a},\Delta T(a,\hat{a})\right)$ and may be solved for analytically through eq. 109 in \cite{Irvin2007} according to eq.~\eqref{eqn: hcw}. For a fixed initial velocity at $a$ given by $\mathbf{v}(a)$ the instantaneous $\Delta V$ required for transfer is then 
\begin{equation}\label{eqn:delV}
    \Delta V(a,\hat{a}) := \|\mathbf{v}_{0}\left(a,\hat{a},\Delta T(a,\hat{a})\right) - \mathbf{v}(a)\|_{2}.
\end{equation}
Not all points in $\mathcal{F}_{\rm H}$ may be successfully connected via an NMT under TOF eq. \eqref{eqn:LL_proxy}. As such, the space of waypoints $\cA$ must be constructed to ensure well-posedness of transfers. This can be done by assessing stability of the relevant coefficient matrix in eq. 113 of \cite{Irvin2007}. $\Delta V$ usage in eq. \eqref{eqn:delV} represents a target impulsive burn for high-level planning wherein the LL planner described by Problem~\ref{prob: OptimizatonLLC} is responsible for calculating the continuously defined control policy $u_{i}(t)$ used to achieve it.

\subsubsection{Waypoint Sequence}\label{subsec: waypoint sequence}
The \textit{waypoint sequence} strategy is characterized as an ordered collection of waypoint targets for a vehicle to sequentially transition between. In this, $\cA := \{a_{1},\dots,a_{m}\}$ is composed of points specified in $\mathcal{F}_{\rm H}$ with a strategy taking the form $\{a_{k}\}_{k=1}^{K}$ with $a_{k}\in\mathcal{A}$ for all $k=1,\dots K$. For such specification, the inspection vehicle follows guidance according to $a_{1}\mapsto a_{2}; a_{2}\mapsto a_{3},\dots, a_{K-1}\mapsto a_{K}$.  For an example, please see the top-left and top-middle of Figure~\ref{fig: WaypointStrat}.

\subsubsection{Waypoint Hold}\label{subsec: waypoint hold}
The waypoint hold guidance category is a specialization of strategies contained in \ref{subsec: waypoint sequence}, where each waypoint pair in the sequence is identical and $\{a_{k}\}_{k=1}^{K}$ with $a_{k} = a\in\mathcal{A}$ for all $k=1,\dots, K$. This is also referred to as a "Hill-static Point Hold", "Waypoint Parking" or simply "Point Hold". For an example, please see top-left and top-right panel of Figure~\ref{fig: WaypointStrat}.

\subsubsection{NMC Hold}\label{subsec: NMC Hold}
\begin{figure}[h!]
\centering
\begin{subfigure}{0.33\linewidth}
\includegraphics[width=1.2\linewidth]{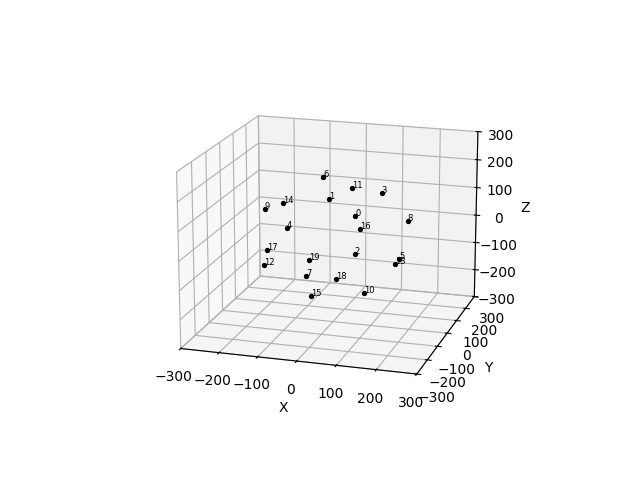}
\end{subfigure}
\begin{subfigure}{0.33\linewidth}
\includegraphics[width=1.2\linewidth]{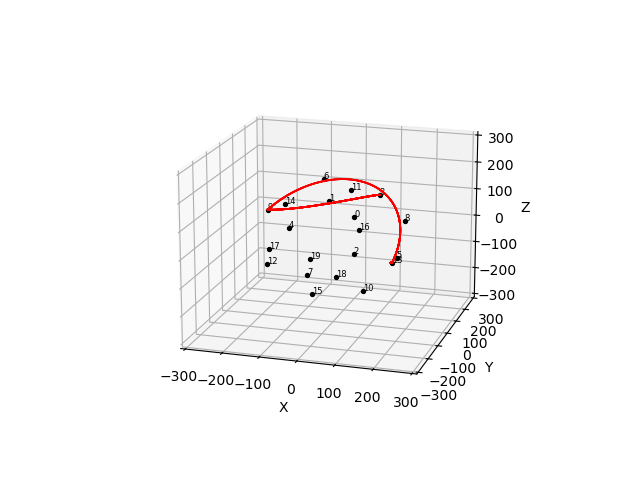}
\end{subfigure}
\begin{subfigure}{0.33\linewidth}
\includegraphics[width=1.2\linewidth]{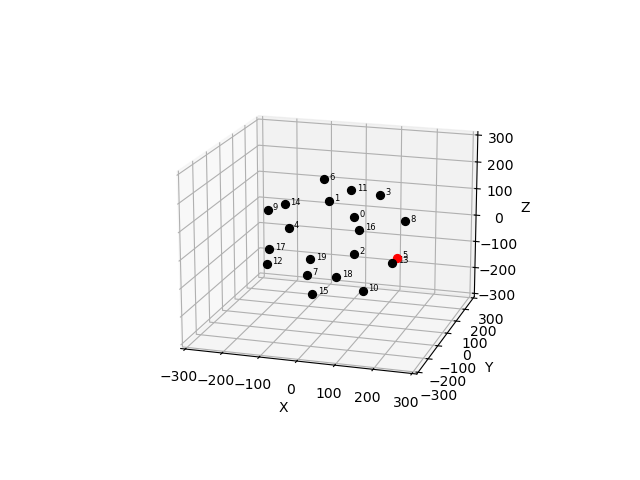}
\end{subfigure}
\begin{subfigure}{0.33\linewidth}
\includegraphics[width=1.2\linewidth]{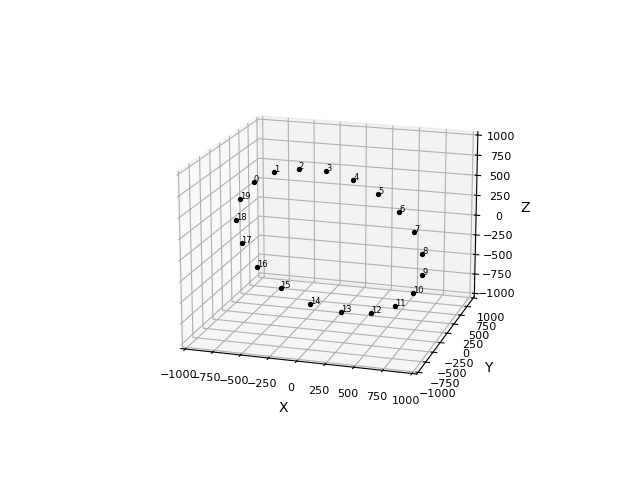}
\end{subfigure}
\begin{subfigure}{0.33\linewidth}
\includegraphics[width=1.2\linewidth]{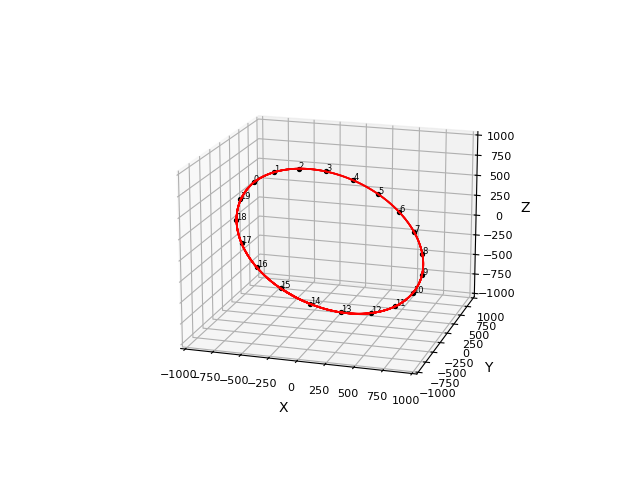}
\end{subfigure}
\caption{The top-left figure demonstrates an action space specified for waypoint sequence and waypoint hold inspection strategy. The top-middle figure demonstrates an example trajectory corresponding to a waypoint sequence strategy; the top-right figure shows an example trajectory corresponding to a point-hold. The bottom panel demonstrates an example action specification for an NMC hold.}
\label{fig: WaypointStrat}
\end{figure}

In this specification, actions take the form of translationally discretized NMCs spanning a full orbital period. The elements $a_{i}\in\mathcal{A}$ are defined such that $a_{i} = \{a_{i,j}\}_{j=1}^{\overline{m}}$ under which there exists a time partition $t_{i} = \{t_{i,k})\}_{k=1}^{\overline{m}}$ satisfying 
\begin{equation}\label{eqn: NMC}
    a_{i,0} = a_{i,\overline{m}},\quad (t_{i,\overline{m}} - t_{i,0}) = \frac{2\pi}{\sigma_{\mu}},\, \text{ and }a_{i,k} = f_{\text{Hill}}(t_{i,k}, 0),\text{ for } k = 1,\dots,\overline{m},\nonumber.
\end{equation}
A sufficient condition in the generation of actions satisfying \eqref{eqn: NMC} can be referenced above in \eqref{eqn: NMC_init}. For an example action and corresponding vehicle trajectory, see the bottom panel of Figure~\ref{fig: WaypointStrat}.

\subsection{Problem Definition}
The formulation used to structure our analysis is comprised of a rotationally dynamic RSO, $n$ inspecting agents, and a flexible notion of control actions in concurrence with the waypoint sequence, point hold, and NMC hold strategies defined above. This is provided in a package that has capacity to support various levels of environmental fidelity. A matrix of modeling factors used for internal and external strategy validation is described below in Table~\ref{tab:ValidationFactors}. Key features of the strategy space that will be explored includes passivity with respect to actuation frequency - see \ref{subsec: NMC Hold},  passivity with respect to translational movement - see \ref{subsec: waypoint hold}, and an active strategy which requires frequent actuation and translation target updates - see \ref{subsec: waypoint sequence}.

Intuitively, each action $a\in\cA$ gives potential opportunities to retrieve an image of the target under a frame centered attitude pointing command at viewpoints $v\in\mathcal{F}_{\rm H}$ determined by $a$. The inspection target is assumed to be a rigid body described by a set $\cP = \{p_{1},\dots,p_{N}\}$ of points of interest (POIs) distributed along its surface. The evolution of attitude dynamics is modeled by Euler's rotational equations of motion for rigid bodies. Supposing that the body frame $\mathcal{F}_{\rm B}$ of the RSO aligns with the principal axes of inertia and that there aren't externally applied torques, rotational equations of motion takes the following form: 
\begin{align}
    I_{xx}\dot{\omega}_x &= (I_{yy} - I_{zz})\omega_y\omega_z\nonumber \\
    I_{yy}\dot{\omega}_y &= (I_{zz} - I_{xx})\omega_x\omega_z\label{eq:ERQ_MOI_NoTorque}\\
    I_{zz}\dot{\omega}_z &= (I_{xx}- I_{yy})\omega_x\omega_y\nonumber 
\end{align}
where $I \in \mathbb{R}^{3\times3} = \text{Diag}(I_{xx}, I_{yy}, I_{zz})$ is the inertia matrix, and $\mathbf{\omega} \in \mathbb{R}^{3} = [\omega_x \ \omega_y \ \omega_z]^T$ is the angular velocity of the inspection target in $\mathcal{F}_{\rm E}$; see \cite{Peter04}. As in \cite{FMark}, we may then recover a quaternion representation $q^{BF}_{ECI}$ of attitude dynamics from $\mathcal{F}_{\rm B}$ to $\mathcal{F}_{\rm E}$ by solving eq.~\eqref{eq:ERQ_MOI_NoTorque}. This is given by:
\begin{align}
    \dot{q}^{BF}_{ECI} =
    \frac{1}{2}
    q^{BF}_{ECI}
    \otimes[0,\mathbf{\omega}]^{T}
    .
    \label{eq:ERQ_att}
\end{align}

The set of POIs visible in a single image taken from viewpoint $v$ (defined by an action $a\in\mathcal{A}$) within a $\theta^\circ$ field of view (FOV) is determined by:
\begin{equation}\label{eqn:point_visibility}
    \mathbf{p}(v) := \left\{p\in\cP_{v} \big| \arccos{\left(\frac{v\cdot p}{\|v\|\| p\|}\right)}\ge \theta^\circ \right\}
\end{equation}
where $\cP_{v_{i}}\subset \cP$ is the set of points that are illuminated with an acceptable intensity and aren't self occluded from $v_{i}$. The point occlusion operation to generate $\cP_{v_{i}}$ is handled through hidden point removal defined in \cite{Katz07}. 

For an underlying state space $\mathcal{S}$ defined by deputy translational state, sun position, RSO rotational state, and POI observability, the inspection problem itself is described below in Problem~\ref{prob:VSP}.
\begin{problem}[Inspection]\label{prob:VSP}
     Specified for an agent-factored transition and observation independent decentralized, partially observable Markov decision process (DEC-POMDP), each deputy aims to find a policy $\pi:\mathcal{S}\mapsto\mathcal{A}$ satisfying:
    \begin{align}\label{eqn:agentPol}
        \pi_{i}^{*} &\in \argmax\limits_{\pi_{i}}\sum\limits_{a\in\mathcal{A}}\pi_{i}(a|\overline{o}_{i})Q_{i}^{\pi_{i}}(\overline{o}_{i},a),\text{ where } \\ 
        Q_{i}^{\pi_{i}}(\overline{o}_{i},a_{i})&:=\E^{\pi}\left[\sum_{k=0}^{T}\gamma^{k} \left(\frac{|\cup_{\overline{v}}\mathbf{p}(v)|}{|\cP|} - \Delta V(a_{i},a_{k})\right)\big| \overline{O}_{i,t}=\overline{o}_{i},\,a = a_{i}\right]\nonumber
    \end{align}
for each $i=1,2,\dots, n$ with $\overline{o}_{i}$ describing agent observability of state $\mathcal{S}$ and $\overline{v}$ represents an induced sequence of viewpoint locations determined by a fixed imaging rate along nominal reference trajectories determined by the specific HL guidance category. $T\in\mathbb{R}_{+}$ is the planning time horizon.
\end{problem}
It should be noted that the expectation in Problem~\ref{prob:VSP} is taken against uncertainty propagated via initial state $s_{0}\in\mathcal{S}$. At each simulated time step, a low-level controller is passed an observation tuple consisting of agent position, velocity, goal position and expected traversal time.  While high-level actions are informed by a fuel penalty calculated through instantaneous burn, low-level actions are taken in accordance with a fixed burn-rate approximating continuous thrust inputs $u(t)$ necessary to produce the instantaneous burn expected by the high-level planner. Notationally, we distinguish between two different time scales (high vs low level) using the following notation: $\delta r_{i}(t)$ denotes agent position in the Hill frame at time $t$, for a feasible time interval $(s-t)>0$ we denote $\mathbf{v}_{0}(\delta r_{i}(t),\delta r_{i}(s))$ as the \emph{initial velocity} required to traverse between $\delta r_{i}(t)$ and $\delta r_{i}(s)$ over $(t-s)$ TOF whereas $\mathbf{v}_{f}(\delta r_{i}(t),\delta r_{i}(s))$ represents the \textit{final velocity}.  To help streamline notation, agent indices will be suppressed during the following exposition.  The optimization problem solved between high-level environment steps is formulated as:
\begin{problem}[Low-Level Planning] \label{prob: OptimizatonLLC}
Fix a starting time $t_{0}\ge0$, starting position $ r_{0}\in\cA$, target traversal time $t_{f}>t_{0}$, and target waypoint position $ r_{f}\in\cA$. The initial agent positional state is given by $\delta r(t_{0})= r_{0}$ and the desired agent control trajectory must satisfy $\delta r(t_{f})= r_{f}$. For a fixed $N$-step discretization of the TOF $\Delta T:=t_{f}-t_{0}$ with uniform mesh width $\Delta t := \frac{\Delta T}{T}$, the low-level planner solves:
\begin{align*}
\min\limits_{\Tilde{\mathbf{u}}}\sum_{j=0}^{n-1}\|\Tilde{\mathbf{u}}(t_{j})\|_{2}\Delta t,\,\text{ s.t. } \|\mathbf{v}_{0}(\Tilde{ r}(t_{j+1}), r_{f})-\mathbf{v}_{f}(\Tilde{ r}(t_{j}),\Tilde{ r}(t_{j+1}))\|_{2}=0,\quad \forall j\in\{0,\dots,n-1\}
\end{align*}
where $\Tilde{ r}(t_{j})$ is the agent's position at time $t_{j}$ under historical thrust control policy $\Tilde{u}(t_{0}),\dots,\Tilde{u}(t_{j})$ evolving according to a discretization of eq.~\ref{eqn: hcw}.
\end{problem}

Note that since the equality between steps in the constraint is binding, the optimization procedure is implicitly minimizing the variable $\Tilde{r}$ through expected positional drift  away from the unique reference trajectory generated by the high-level planner's target initial velocity. The structure of Problem~\ref{prob:VSP} is simply posed to allow for various refinements to be effectively tested against one another - Problem~\ref{prob: OptimizatonLLC} then permits an additional source of external policy validation to help quantify scenarios that may be sensitive to different low-level control schemes. 

\section{Approach}
Three inspection strategies will be compared in this study. This includes Point Hold (PH) representing \ref{subsec: waypoint hold}, NMC Hold (NMC) representing \ref{subsec: NMC Hold}, and Reinforcement Learning (RL) representing \ref{subsec: waypoint sequence}. The PH and NMC strategies are generated using Monte Carlo analysis conducted in a low-fidelity simulation environment - see "Monte Carlo Simulation" in Table~\ref{tab:ValidationFactors}; the RL strategy is trained in a medium-fidelity simulation environment - see "RL Training Environment" in Table~\ref{tab:ValidationFactors}. Performance will be internally validated prior to comparison and is conducted within each policy's respective simulation environment. A separate validation environment is constructed and includes a set of environmental factors unifying complexity between the Monte Carlo simulation and RL training environment while simultaneously adding features that weren't considered during internal validation; see "Validation Environment" in Table~\ref{tab:ValidationFactors}. RSO translational motion is fixed across all environments with $r_{0}(0) = [7357000\text{m},0,0,0,7360.69\text{m/s},0]$. The nominal distance targeted from agent waypoint to $\mathcal{O}_{\rm H}$ is 200m; only three agent configurations ($n=3$) are considered. Environment specific setup and additional detail is described below. 

\begin{table}[h!]
  \begin{center}
    \begin{tabular}{lcccc} 
    \toprule
     Attributes &  Monte Carlo Simulation & RL Training Environment & Validation Environment \\
      \midrule
      \cellcolor{gray!10}{Time Horizon} & \cellcolor{gray!10}{Fixed} & \cellcolor{gray!10}{Random} & \cellcolor{gray!10}{Random}\\
      Illumination & Binary & Binary & Blinn-Phong and Earth Shadow\\
      \cellcolor{gray!10}{RSO geometry} & \cellcolor{gray!10}{Sphere} & \cellcolor{gray!10}{General - ply file} & \cellcolor{gray!10}{General - ply file} \\
      RSO rotation & Variable and Noisy & Fixed & Variable and Noisy \\
      \cellcolor{gray!10}{RSO translation} & \cellcolor{gray!10}{Fixed, CWH} & \cellcolor{gray!10}{Fixed, CWH} & \cellcolor{gray!10}{Fixed, CWH}\\
      Imaging Rate & Fixed & Fixed & Fixed\\
      \cellcolor{gray!10}{Agent Entrance} & \cellcolor{gray!10}{Unmodeled} & \cellcolor{gray!10}{Uncertain} & \cellcolor{gray!10}{Uncertain}\\
      \bottomrule
    \end{tabular}
  \end{center}
  \caption{Table specifying main factors considered for internal and eternal validation. Each column specifies key modeling assumptions associated with each testing environment. The validation environment includes higher-fidelity versions of the features used to generate both active and passive policies. A fixed time horizon and imaging rate indicates predetermined, deterministic values used during simulation. Binary illumination treats differing illumination intensities as equal.}
  \label{tab:ValidationFactors}
  \vspace{-15pt}
\end{table}

\subsection{Monte Carlo Sampling}

The Monte Carlo simulation environment is built from the work in \cite{SC2024}. It consists of three major components: a randomization of uncertain rigid body dynamics, waypoint or NMC based agent trajectory specification, and a camera model that includes field of view, occlusion, and illumination. Taken together, it allows for effective performance evaluation of different deputy actions under various RSO behavior and environmental assumptions. This environment leverages random sampling of $s_{0}\in\mathcal{S}$ to explore the impact different assumptions can have on inspection performance. To accurately reflect sensitivity of performance to uncertainty in $s_{0}$, a brute force approach is used to fully cover sampling of different parameter spaces. Although this approach is computationally burdensome and scales poorly with environmental complexity, it can also provide great insight into the utilization of \textit{simple} inspection strategies to solve Problem~\ref{prob:VSP}. One of the key advantages of this is in predictability and transferability in the strategy space - specifically being able to provide the complete inspection strategy and provide ad-hoc safety analysis prior to the inspection mission. 

Importantly, the Monte Carlo environment assumes RSO geometry is spherical; all Monte Carlo samples are forward propagated under eq.~\eqref{eqn: hcw}. Actions are passively defined and are drawn from a discrete set of twenty options (either type \ref{subsec: waypoint hold} or \ref{subsec: NMC Hold}). Fuel calculation is performed only for action type \ref{subsec: waypoint hold}. Deputies of action type \ref{subsec: NMC Hold} are assumed to incur zero fuel cost for maintaining the NMC based action. Based on the random samples above, Different combinations of agent actions are explored and maximized to provide candidate solutions to Problem~\ref{prob:VSP}. Low level vehicle control and deputy entrance conditions are unmodeled in the Monte-Carlo environment. The simulation environment from \cite{SC2024} was setup under configuration parameters for both PH and NMC strategies as seen below in Table~\ref{tab:MCconfig}.

\begin{table}[h!]
  \begin{center}
    \begin{tabular}{lccccccccccc} 
    \toprule
     RSO Mode & N & $\omega (\text{rad/s})$ & $I$ & $q_{nom}$ & b & $\theta_{u}$ & dt (s)& $n_{key}$ & $n_{agloc}$ & $\theta_{FOV}$ & $\theta_{OCC}$\\
      \midrule
      \cellcolor{gray!10}{Static CWH} & \cellcolor{gray!10}{1000} & \cellcolor{gray!10}{$\begin{bmatrix}0 \\0 \\ \sigma_{\mu}\end{bmatrix}$}& \cellcolor{gray!10}{$\begin{bmatrix}100 \\70 \\50\end{bmatrix}$}  & \cellcolor{gray!10}{$\begin{bmatrix}1\\0 \\0\\0\end{bmatrix}$} & \cellcolor{gray!10}{20\%} & \cellcolor{gray!10}{$\frac{\pi}{6}$} & \cellcolor{gray!10}{1s} & \cellcolor{gray!10}{20} & \cellcolor{gray!10}{20} & \cellcolor{gray!10}{$\frac{\pi}{6}$} & \cellcolor{gray!10}{$\frac{\pi}{6}$}\\
      Stable Tumble & 1000 & $\begin{bmatrix}5\sigma_{\mu} \\0 \\ 50\sigma_{\mu}\end{bmatrix}$& $\begin{bmatrix}100 \\70 \\50\end{bmatrix}$ & $\begin{bmatrix}1\\0 \\0\\0\end{bmatrix}$&20\%&$\frac{\pi}{6}$&1s&20&20&$\frac{\pi}{6}$&$\frac{\pi}{6}$\\
      \bottomrule
    \end{tabular}
  \end{center}
  \caption{Configuration used for Monte Carlo simulation of both PH and NMC strategies. All parameters are specified according to the notation in \cite{SC2024}. Units of $\theta_{FOV}, \theta_{OCC}$ and $\theta_{u}$ are specified in radians.}
  \label{tab:MCconfig}
  \vspace{-15pt}
\end{table}

Results are aggregated by strategy type and averaged between the two distinct RSO dynamic modes. The Pareto front is calculated across all simulation results optimizing average inspection percentage vs. time. The best performing strategies can be seen below in Figure~\ref{fig: MC_Rollout}.

\begin{figure}[h!]
\centering
\begin{subfigure}{0.95\linewidth}
\includegraphics[width=1\linewidth]{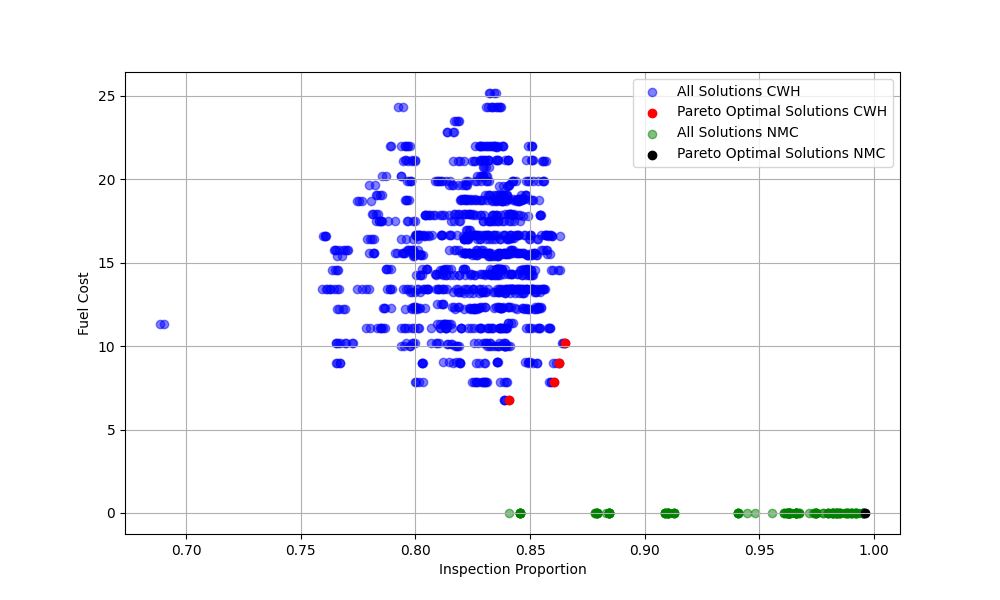}
\end{subfigure}
\begin{subfigure}{0.48\linewidth}
\includegraphics[width=1.1\linewidth]{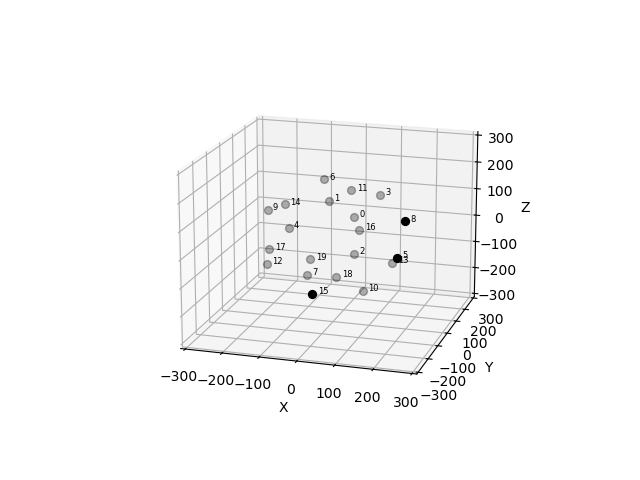}
\end{subfigure}
\begin{subfigure}{0.48\linewidth}
\includegraphics[width=1.1\linewidth]{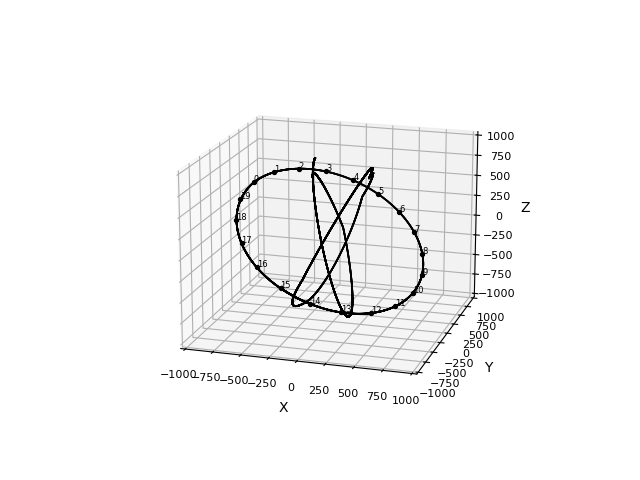}
\end{subfigure}
\caption{The figure in the upper panel contains the MonteCarlo results used for policy determination of PH and NMC strategies. The Pareto front for PH is highlighted in red, whereas results for NMC are highlighted black. The bottom left figure contains the strategy selected for PH whereas the bottom right figure contains the strategy selected for NMC.}
\label{fig: MC_Rollout}
\end{figure}

\subsection{RL Training Environment}\label{subsec: RL Training Env}

The multiagent reinforcement-learning (RL) based controller used is detailed in \cite{JSR24}. This is specified to generate a waypoint sequence in accordance with \ref{subsec: waypoint sequence}. In contrast to the Monte Carlo approach, the RL training approach excels at navigating scenarios in more complex environments. Policies can be specified to be centrally or decentrally coordinated and have support for both discrete and continuous actions. This is notoriously difficult as the problem being solved has a solution space whose complexity scales nondeterministic super-exponentially with agent number. Different RL training methods have shown great promise in converging on solutions  exhibiting exceptional performance. However, this can easily come at the cost of model overfitting wherein the RL policy will not reliably transfer to new scenarios. In contrast to the passively defined PH and NMC strategies above, the waypoint sequence is generated as a response to current system state. This significantly complicates maneuver predictability and corresponding prehoc safety analysis. 

\begin{figure}[h!]
\centering
\begin{subfigure}{0.45\linewidth}
\includegraphics[width=1\linewidth]{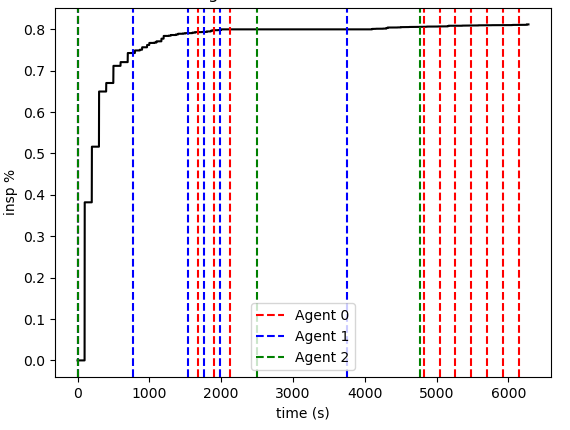}
\end{subfigure}
\begin{subfigure}{0.45\linewidth}
\includegraphics[width=1.08\linewidth]{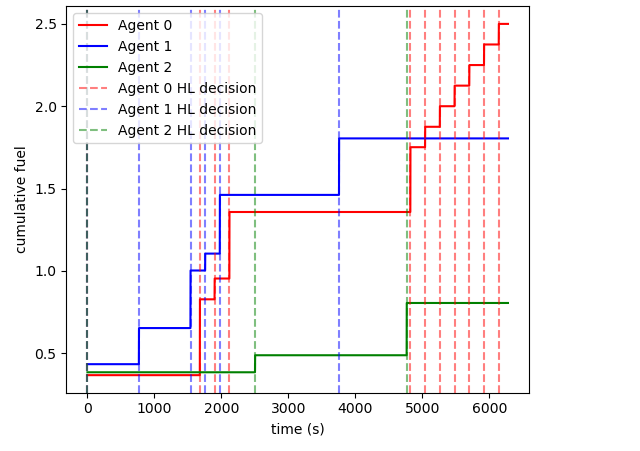}
\end{subfigure}
\begin{subfigure}{0.55\linewidth}
\includegraphics[width=1\linewidth]{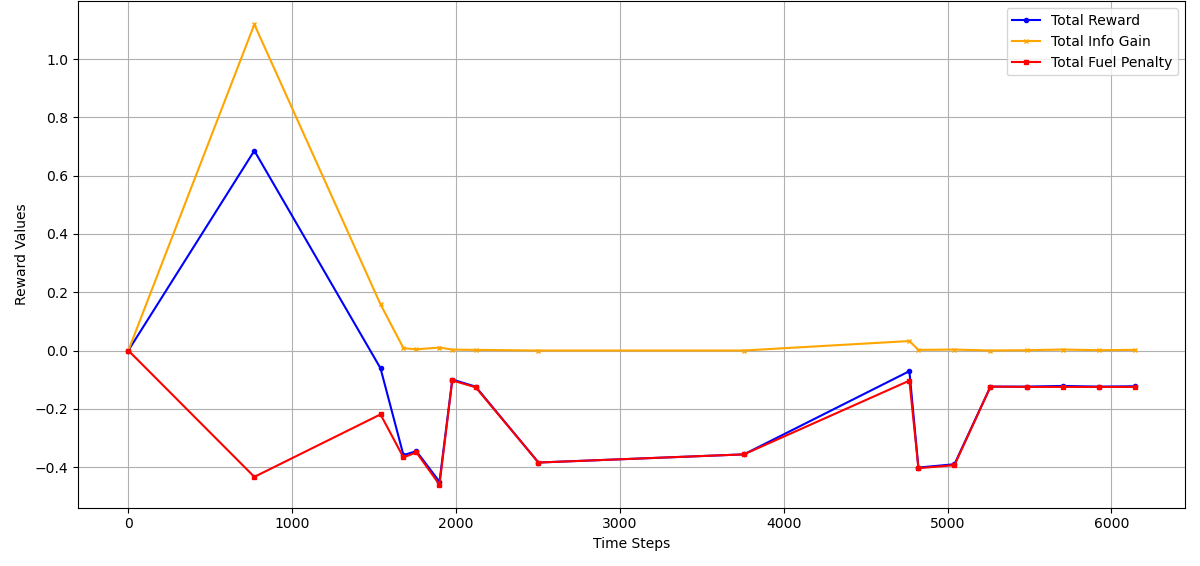}
\end{subfigure}
\begin{subfigure}{0.4\linewidth}
\includegraphics[width=1\linewidth]{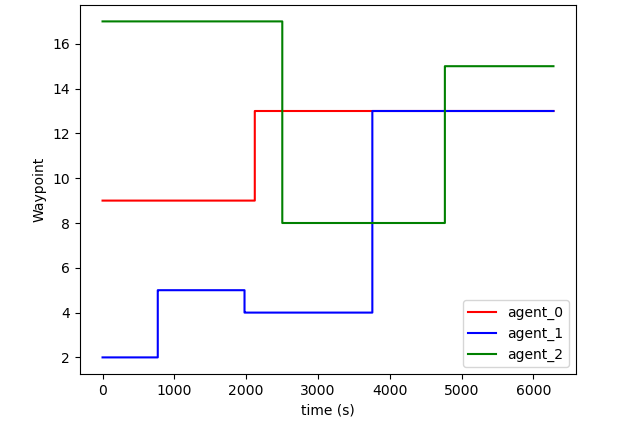}
\end{subfigure}
\caption{Sample policy rollout with overlayed HL decisions generated by the trained RL policy used for comparisons.}
\label{fig: RL_Rollout}
\end{figure}

Primary assumptions made in this environment include: there is a known geometric primitive of the RSO for training, fuel cost for actions is described by eq.~\eqref{eqn:delV}, image taking is coupled with waypoint transfer, the RSO rotational state is perfectly observable, and peer translational state is perfectly observable. This was generalized to add support for: fixed imaging rate that is decoupled from waypoint transfer, RSO illumination using a generalized Blinn-Phong model, and asynchronous decision making. The policy training process is conducted by sampling the action-value function corresponding to an objective specified through a proxy objective known as a reward function. This provides a measurement of utility relating different actions back to the primitive objective in Problem~\ref{prob:VSP}. The information accrued during this sampling is managed by a training algorithm - for this work, we used \textit{Recursive Replay Distributed DQN (R2D2)}; see \cite{Mnih13, Mnih15}. An example rollout of the trained RL policy is shown in Figure~\ref{fig: RL_Rollout}.

\begin{table}[h!]
  \begin{center}
    \begin{tabular}{lcccccccc} 
        \toprule
         Hyperparameter: & $\alpha$ & $\beta$ & $r_{0}$ & M & LR & BS & DR & PE \\
         \midrule
         \cellcolor{gray!10}{Value:} & \cellcolor{gray!10}{2} & \cellcolor{gray!10}{1} & \cellcolor{gray!10}{0} & \cellcolor{gray!10}{.85} & \cellcolor{gray!10}{5e-4} & \cellcolor{gray!10}{256} & \cellcolor{gray!10}{.95} & \cellcolor{gray!10}{.6}\\
         \midrule
         Sim Config: & Geometry & $\omega$ & $I$ & $q_{nom}$ & $U_{r}$ & $\Delta t$ & $\overline{t}$ & \\
         \midrule
         \cellcolor{gray!10}{Value:} & \cellcolor{gray!10}{Aura} & \cellcolor{gray!10}{$\left(\begin{bmatrix}5\sigma_{\mu} \\0 \\ 50\sigma_{\mu}\end{bmatrix}, \begin{bmatrix}0 \\0 \\ \sigma_{\mu}\end{bmatrix}\right)$} & \cellcolor{gray!10}{$\begin{bmatrix}100 \\70 \\50\end{bmatrix}$} & \cellcolor{gray!10}{$\begin{bmatrix}1\\0 \\0\\0\end{bmatrix}$} & \cellcolor{gray!10}{Unif} & \cellcolor{gray!10}{1 s}&\cellcolor{gray!10}{100 s} &\cellcolor{gray!10}{} \\
        \bottomrule
    \end{tabular}
  \end{center}
  \caption{Configuration used for training of Waypoint Sequence/RL strategy. LR: Learning Rate, BS: Batch Size, DR: Discount Rate, PE: Priority Exponent, $U_{r}$: Sampling distribution for initial deputy translational state, $\Delta t$: simulation time step, $\overline{t}$: deputy imaging rate; all other parameters are specified according to the notation in \cite{JSR24}. The Aura satellite was used for RSO geometry, retrieved from https://nasa3d.arc.nasa.gov.}
  \label{tab:RLconfig}
  \vspace{-15pt}
\end{table}

\subsection{Validation Environment}
\begin{figure}[!htb]
\centering
\includegraphics[width=.8\linewidth]{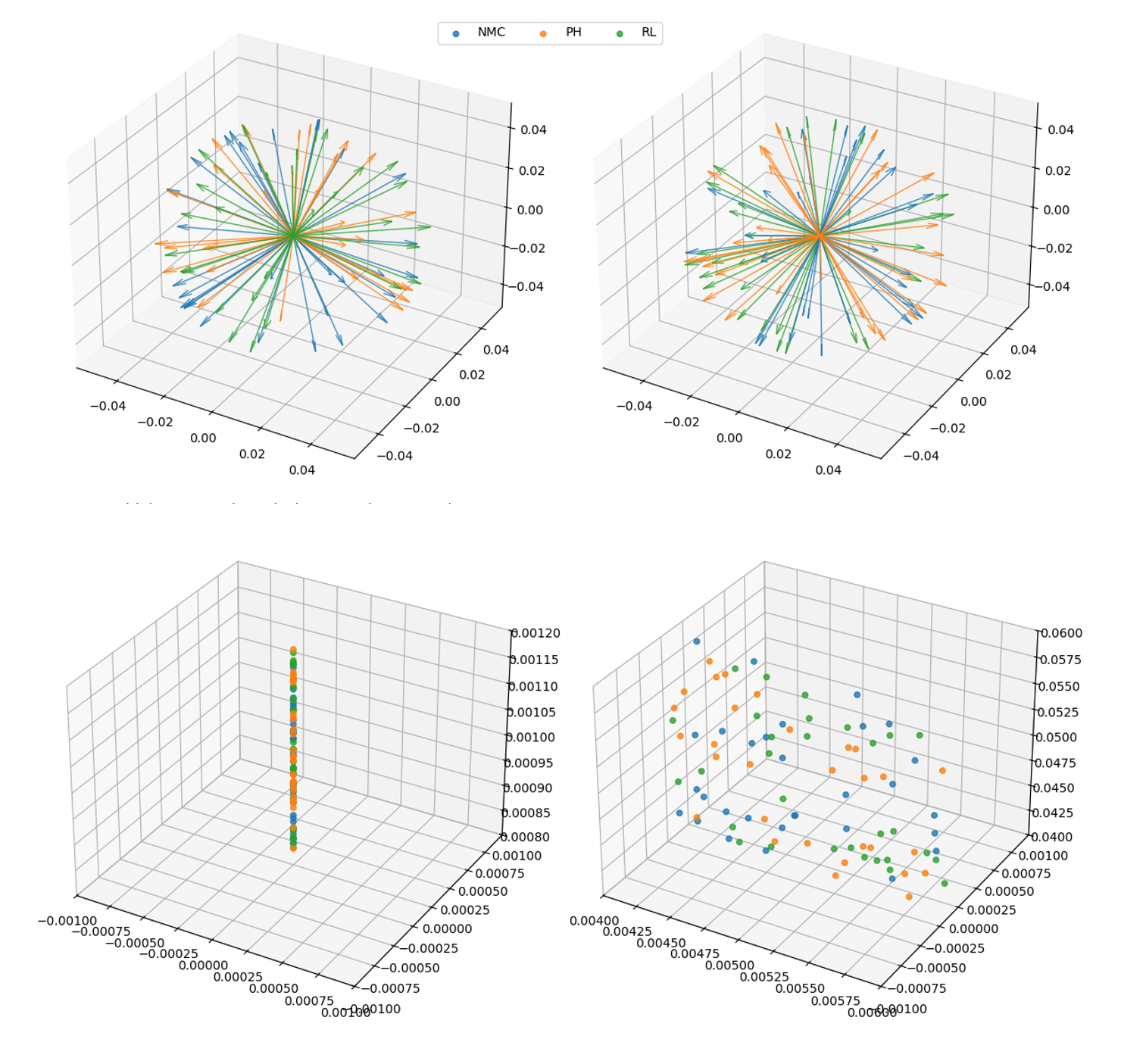}
\caption{The top panel shows sampling of RSO attitude over the 30 trials conducted for each RSO dynamic mode. The bottom panel demonstrates initial angular velocity sampled for the same trials.}
\label{fig: initState}
\end{figure}

Policy performance will be assessed in the validation environment according to two metrics directly related to the objective function in Problem~\ref{prob:VSP}: fuel usage and inspection percent. Timing considerations for mission completion are implicitly encoded through episode time horizon, specified by: 
\begin{equation}\label{eqn:timeHorizon}
	\tau:=\inf\left\{t\ge 0\,,\,\frac{\|\mathbf{p}(t)\|_{1}}{|\cP|}\ge M\right\}
\end{equation}
for a fixed threshold $M\in[0,1]$. If $\tau \le T$, the inspection mission is successfully accomplished. If $\tau > T$, the mission is designated a failure. Once $\tau$ is triggered, the instance is terminated and performance metrics are no longer calculated. This ties duration of the mission directly to primitive performance metrics allowing for statistical analysis to be conducted on a smaller testing factor matrix, as opposed to including success time directly as a variable. The validation environment is an augmented form of the training environment used in Section~\ref{subsec: RL Training Env}; it contains higher fidelity illumination and adds uncertainty to the RSO rotational dynamics. The basis for performance validation is in random sampling from comparatively uncertain and information denied scenarios. Uncertainty itself is propagated through initial RSO attitude, RSO angular velocity (samnpled around Static CWH and Stable tumble dynamic modes; see Table~\ref{tab:MCconfig}), and initial deputy translational state. It is important to note that the PH and NMC policies were generated in environments where agent entrance criteria (initial deputy translational state) were unmodeled. As such, those strategies will see an increased fuel penalty and \textit{decreased} inspection performance due to an unplanned entrance maneuver prior to the planned strategy taking hold. Meanwhile, the RL policy was not trained under uncertainty propagated in RSO rotational dynamics. This has a direct impact on planning performance where implicit hidden state estimation can be sensitive due to the complexity of RSO surface geometry. Figure~\ref{fig: initState} provides an illustration of all seeding initial RSO rotational states used for policy comparison. 

Illumination adds a Blinn-Phong filter and also incorporates a region where the RSO is in Earth's shadow - preventing any an all information from being solicited. This has a huge impact on solvability, where opportunity sets may drastically shrink. All three policies (PH, NMC, and RL) were generated using a binary ``on/off" style of illumination without Earth shadow as a consideration. It is intuitive that the RL policy has a chance to minimize the impact by actively choosing low-fuel waypoints during this period of time - but it has not been trained to do so. Conversely, the passively defined PH and NMC based strategies cannot directly mitigate the effects and represent a baseline for the degree of care we may need to take when planning. This is designed to help provide insight into strategy selection for future inspection missions. All low-level control is handled by direct numerical solution of Problem~\ref{prob: OptimizatonLLC} specified by the action class: \ref{subsec: NMC Hold}, \ref{subsec: waypoint hold}, or \ref{subsec: waypoint sequence}. Configuration of environmental parameters can be seen below in Table~\ref{tab:Valconfig}. It should be noted, that the process of \textit{internal validation} is conducted in \cite{SC2024} and \cite{JSR24}. \textit{External validation} is not considered in this paper but can be conducted in the same environment as future work.

\begin{table}[h!]
  \begin{center}
    \begin{tabular}{lccccccc} 
        \toprule
         Config: & Illum (RSO) & Illum (Sun) & Capture Criteria & $U_{\omega}$ & $U_{q}$ & T (s) & EA (rads)\\
         \midrule
         \cellcolor{gray!10}{Value:} & \cellcolor{gray!10}{$\begin{bmatrix}0.1\\0.7,\\1\end{bmatrix}$}&\cellcolor{gray!10}{$\begin{bmatrix}1 \\1\\ 1\end{bmatrix}$} & \cellcolor{gray!10}{$\begin{bmatrix}0.9 \\0.1\\ \frac{\pi}{6}\end{bmatrix}$} & \cellcolor{gray!10}{20\%} & \cellcolor{gray!10}{Unif} &\cellcolor{gray!10}{6280} & \cellcolor{gray!10}{$[\frac{2\pi}{3},\frac{4\pi}{3}]$} \\
        \bottomrule
    \end{tabular}
  \end{center}
  \caption{Configuration used in the validation environment for policy comparison. All simulation configuration parameters from Table~\ref{tab:RLconfig} are inherited. The vector for illumination properties contains properties for ambience, diffusivity, and specularity. Capture criteria specifies threshold for acceptance of POI based on brightness, darkness, and FOV. EA is ``Earth Angle" and represents the feasible range of sun angle for which the Earth's shadow blocks all illumination. $U_{\omega}$ and $U_{q}$ represent the sampling scheme for uncertainty in RSO rotational state; the value for $U_{\omega}$ is quoted as a linear proportion of the reference value for $\omega$.}
  \label{tab:Valconfig}
  \vspace{-15pt}
\end{table}

\section{Results}
Results for policy comparison are specified by metric: Fuel Consumption and Inspection Percentage. There were 30 trajectories sampled as a basis for statistical significance.

\subsection{Fuel Consumption}\label{sec: fuel-consumption}
\begin{figure}[!htb]
\centering
\includegraphics[width=.9\linewidth]{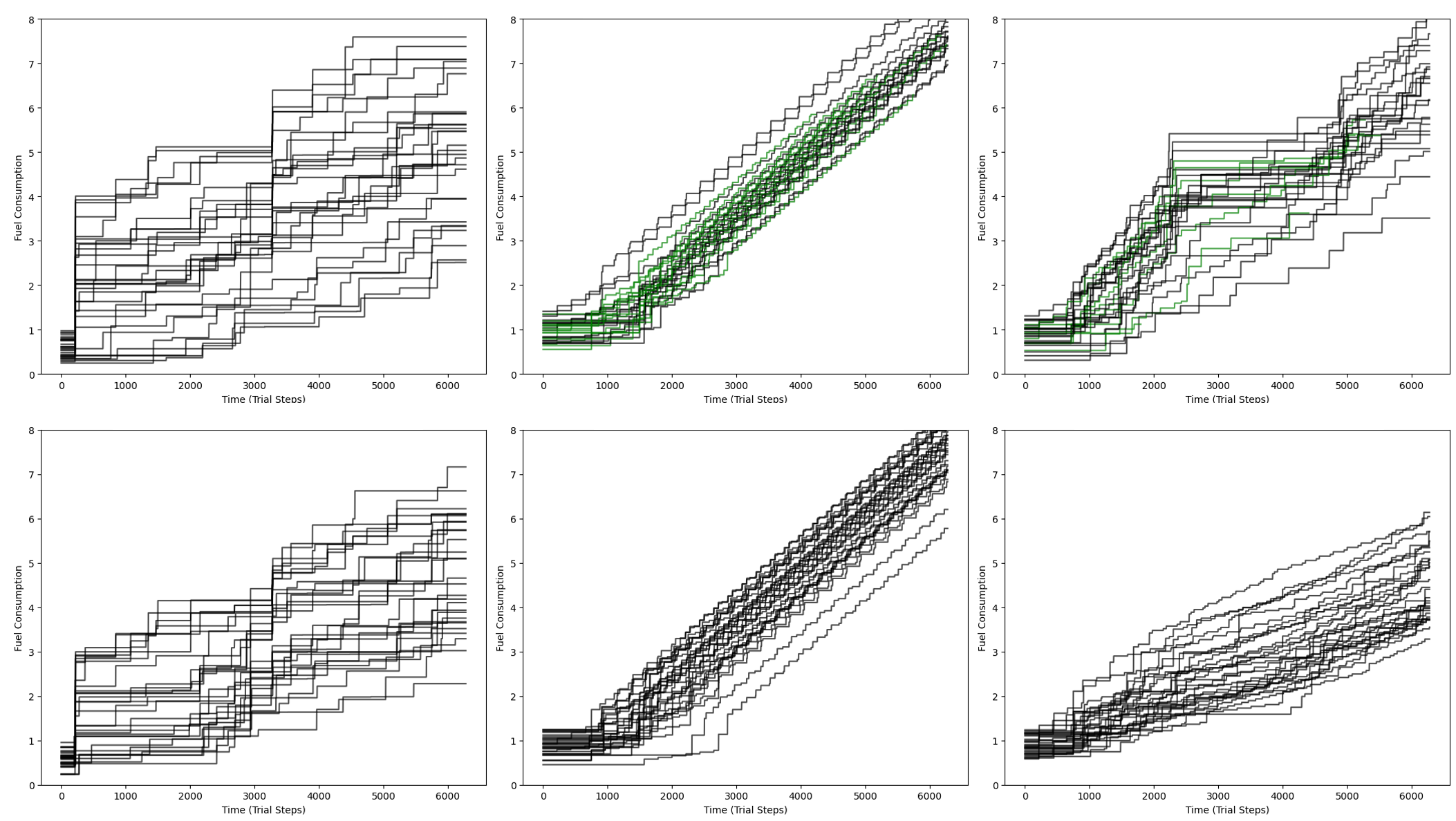}
\caption{Cumulative fuel usage over time for the three agent inspection mission across 30 trials. Specified by RSO dynamic mode and policy combination. Green traces indicate early mission success. The top panel is for the Stable Tumble mode while the bottom is for Static CWH.}
\label{fig: fuel_traj1}
\end{figure}
Fuel consumption for each strategy was greatly determined by time to inspection completion. Summarily, the required entrance condition had a significant impact on fuel usage for NMCs, while PH and RL were better controlled. For the stable tumble RSO dynamic mode, both PH and RL strategies had a higher proportion of early mission completion when compared against the static CWH RSO dynamic mode. This increases the variance in fuel usage expected due to early termination; see Figure~\ref{fig: fuel_traj1}. The PH policy demonstrated the highest fuel consumption, particularly under the static CWH RSO dynamic mode (Mean = 7.554, SD = 0.646). In contrast, the RL policy exhibits comparably lower fuel usage (Mean = 4.536, SD = 0.806); comparable to the NMC policy (Mean = 4.828, SD = 1.241). The stable tumble RSO dynamic mode appeared to punish excessive agent movement. In this respect, fuel usage for PH decreases while fuel usage for NMC and RL increases. For additional detail, please see Table~\ref{tab:unnamed-chunk-3} below.

\begin{table}[!ht]
\centering
\begin{tabular}[t]{lllrrr}
\toprule
Dynamic & Policy & Variable & Sample Size & Mean & Standard Deviation\\
\midrule
\cellcolor{gray!10}{StableTumble} & \cellcolor{gray!10}{NMC} & \cellcolor{gray!10}{Fuel Consumption} & \cellcolor{gray!10}{30} & \cellcolor{gray!10}{5.124} & \cellcolor{gray!10}{1.503}\\
StaticCWH & NMC & Fuel Consumption & 30 & 4.828 & 1.241\\
\cellcolor{gray!10}{StableTumble} & \cellcolor{gray!10}{PH} & \cellcolor{gray!10}{Fuel Consumption} & \cellcolor{gray!10}{30} & \cellcolor{gray!10}{6.828} & \cellcolor{gray!10}{1.597}\\
StaticCWH & PH & Fuel Consumption & 30 & 7.554 & 0.646\\
\cellcolor{gray!10}{StableTumble} & \cellcolor{gray!10}{RL} & \cellcolor{gray!10}{Fuel Consumption} & \cellcolor{gray!10}{30} & \cellcolor{gray!10}{5.680} & \cellcolor{gray!10}{1.649}\\
\addlinespace
StaticCWH & RL & Fuel Consumption & 30 & 4.536 & 0.806\\
\bottomrule
\end{tabular}
\caption{\label{tab:unnamed-chunk-3}Summary statistics for inspection performance across
different combinations of RSO Dynamic Mode and policy.}
\end{table}

\begin{figure}[h!]
\centering
\begin{subfigure}{0.7\linewidth}
\includegraphics[width=1\linewidth]{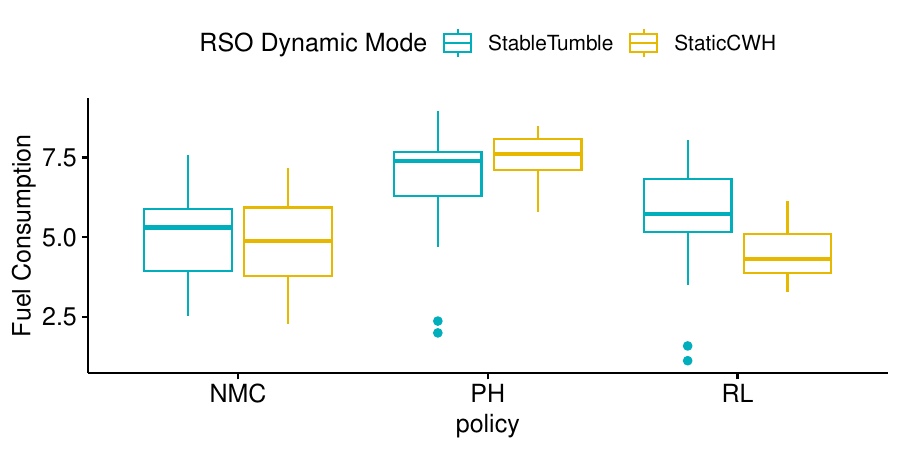}
\end{subfigure}
\begin{subfigure}{0.7\linewidth}
\includegraphics[width=1\linewidth]{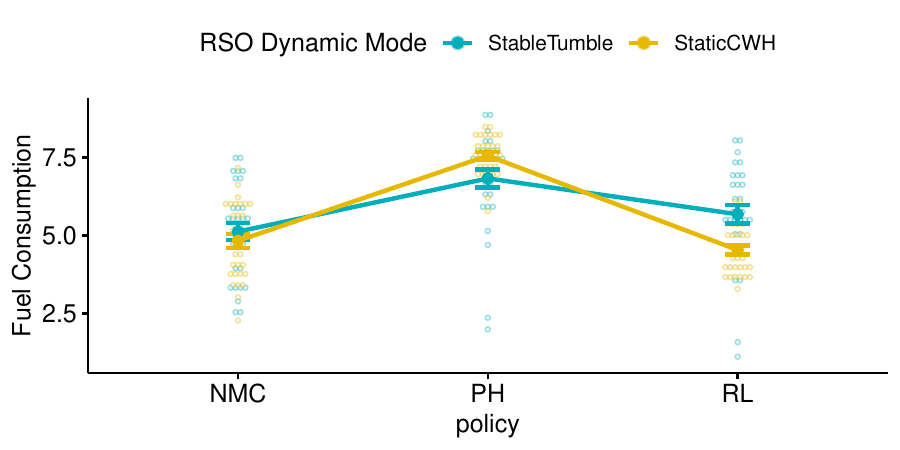}
\end{subfigure}
\caption{The upper panel shows box plot results for fuel consumption versus policy specification. The lower panel shows corresponding interaction effects.}
\label{fig: FuelBP}
\end{figure}

Additional analysis indicates the presence of interaction effects between policy specification and dynamic mode on fuel consumption. This makes the attribution of effects due to a single factor difficult to resolve; for additional detail see Figure~\ref{fig: FuelBP}.

\subsubsection{Two-way ANOVA Test}\label{two-way-anova-test}

\begin{figure}[h!]
  \begin{minipage}[b]{.45\linewidth}
    \centering
    \begin{tabular}[t]{lrr}
    \toprule
    Variable & Statistic & p\\
    \midrule
    \cellcolor{gray!10}{residuals(model)} & \cellcolor{gray!10}{0.9531265} & \cellcolor{gray!10}{1.11e-05}\\
    \bottomrule
    \toprule
    (Df1, Df2) & Statistic & p\\
    \midrule
    \cellcolor{gray!10}{(5, 174)} & \cellcolor{gray!10}{3.426137} & \cellcolor{gray!10}{0.0056034}\\
    \bottomrule
    &&\\
    &&\\
    \end{tabular}
    \captionof{table}{Shapiro (top) and Levene's (bottom) test results conducted on fuel utilization.}\label{tab:ANOVA_assump_fuel}
  \end{minipage}
  \hfill
  \begin{minipage}[b]{.45\linewidth}
    \centering
    \includegraphics[width=0.7\linewidth]{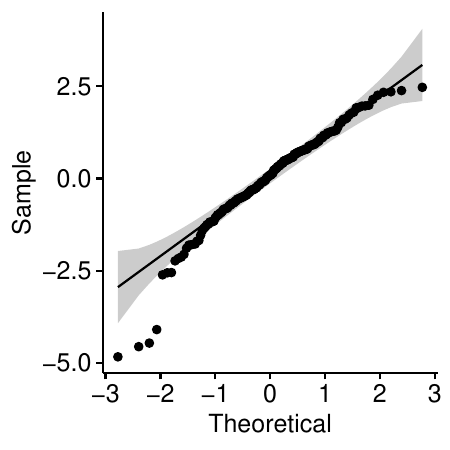}
    \captionof{figure}{Quantile-Quantile (QQ) plot for sampled trials against fuel utilization.}\label{fig: QQplotFuel}
  \end{minipage}
\end{figure}

A Two-Way Analysis of Variance (ANOVA) test was conducted to determine if there is a significant impact of RSO Dynamic Mode and Policy on fuel consumption while also characterizing potential interaction effects. Before beginning, we test the relevant assumptions of normality and homogeneity of variance. Both the significant p-value in a Shapiro test (Table~\ref{tab:ANOVA_assump_fuel}) and the QQ plot show that the assumption of normality is not met; see Figure~\ref{fig: QQplotFuel}. Furthermore, Levene's test shows that there is a significant difference in variance across the groups, indicating that the assumption of homogeneity of variances has not been met; see Table~\ref{tab:ANOVA_assump_fuel}.

Included for completeness, the analysis of variance on fuel consumption showed a statistically significant interaction effect between policy and dynamic on fuel consumption, with an F-ratio of 7.7843 under degrees of freedom 2 and 174, and a p-value of \ensuremath{6\times 10^{-4}}. Since the interaction effect is significant, the policy effect cannot be generalized. This suggests that both the policy specification and the particular RSO dyamic mode have a significant impact on fuel utilization. This is not unexpected as some inspection scenarios can be accomplished much more rapidly than others with corresponding impacts to fuel savings.

\begin{table}[!ht]
\centering
\begin{tabular}[t]{lrrrrr}
\toprule
  & Df & Sum Sq & Mean Sq & F value & Pr(>F)\\
\midrule
\cellcolor{gray!10}{policy} & \cellcolor{gray!10}{2} & \cellcolor{gray!10}{185.248459} & \cellcolor{gray!10}{92.624230} & \cellcolor{gray!10}{54.824215} & \cellcolor{gray!10}{0.0000000}\\
dynamic & 1 & 2.544135 & 2.544135 & 1.505872 & 0.2214279\\
\cellcolor{gray!10}{policy:dynamic} & \cellcolor{gray!10}{2} & \cellcolor{gray!10}{26.302716} & \cellcolor{gray!10}{13.151358} & \cellcolor{gray!10}{7.784279} & \cellcolor{gray!10}{0.0005783}\\
Residuals & 174 & 293.968931 & 1.689477 & NA & NA\\
\bottomrule
\end{tabular}
\caption{\label{tab:unnamed-chunk-4}ANOVA Summary for fuel utilization.}
\end{table}

\subsubsection{Aligned Rank Transform}\label{non-parametric-alternative-aligned-rank-transform-art}

The Aligned Rank Transform (ART) procedure is a non-parametric method that allows for the analysis of interaction effects and main effects in a way that is robust to the violations of ANOVA assumptions. This shows a significant interaction effect between policy and RSO dynamic mode; see Table~\ref{tab:unnamed-chunk-8}. We use emmeans (estimated marginal means) analysis with Bonferroni adjustment for pairwise comparisons. This shows significant differences in adjusted mean for the RL policy between the two RSO dynamic modes. This does not extend to either PH or NMC, which is not surprising given the passive nature of both policies. When looking at policy-to-policy comparison, there are significant differences found between PH and the other two policies (NMC and RL) for both dynamic modes. There is no significant difference between NMC and RL for any dynamic mode. For more detail, please see Table~\ref{tab:unnamed-chunk-9}.

Post-hoc pairwise comparisons with a Bonferroni adjustment reveal that the RL policy is significantly different between the two RSO dynamic modes. However, there is insufficient evidence to suggest differences between the other policies (PH and NMC) across the dynamic mode. Additionally, significant differences are observed between PH and the other two policies (NMC and RL) within both RSO dynamic modes, while no significant differences are found between RL and NMC within either mode. More details can be found in Table~\ref{tab:unnamed-chunk-9}.

\begin{table}[!ht]
\centering
\begin{tabular}[t]{lrrrrrr}
\toprule
  & Df & Df.res & Sum Sq & Sum Sq.res & F value & Pr(>F)\\
\midrule
\cellcolor{gray!10}{policy} & \cellcolor{gray!10}{2} & \cellcolor{gray!10}{174} & \cellcolor{gray!10}{214050.63} & \cellcolor{gray!10}{270525.5} & \cellcolor{gray!10}{68.837900} & \cellcolor{gray!10}{0.0000000}\\
dynamic & 1 & 174 & 13868.89 & 470052.3 & 5.133868 & 0.0246956\\
\cellcolor{gray!10}{policy:dynamic} & \cellcolor{gray!10}{2} & \cellcolor{gray!10}{174} & \cellcolor{gray!10}{48776.58} & \cellcolor{gray!10}{435470.9} & \cellcolor{gray!10}{9.744767} & \cellcolor{gray!10}{0.0000974}\\
\bottomrule
\end{tabular}
\caption{\label{tab:unnamed-chunk-8}ANOVA results based on Aligned Rank Transform for fuel utilization.}

\centering
\begin{tabular}[t]{lrrrrr}
\toprule
Contrast & Estimate & SE & Df & t.ratio & p.value\\
\midrule
\cellcolor{gray!10}{NMC-StableTumble, NMC-StaticCWH} & \cellcolor{gray!10}{0.2959314} & \cellcolor{gray!10}{0.3356066} & \cellcolor{gray!10}{174} & \cellcolor{gray!10}{0.8817807} & \cellcolor{gray!10}{1.0000000}\\
PH-StableTumble, PH-StaticCWH & -0.7263027 & 0.3356066 & 174 & -2.1641492 & 0.2863438\\
\cellcolor{gray!10}{RL-StableTumble, RL-StaticCWH} & \cellcolor{gray!10}{1.1436924} & \cellcolor{gray!10}{0.3356066} & \cellcolor{gray!10}{174} & \cellcolor{gray!10}{3.4078367} & \cellcolor{gray!10}{0.0073217}\\
NMC-StableTumble, PH-StableTumble & -1.7037805 & 0.3356066 & 174 & -5.0767197 & 0.0000088\\
\cellcolor{gray!10}{NMC-StableTumble, RL-StableTumble} & \cellcolor{gray!10}{-0.5556843} & \cellcolor{gray!10}{0.3356066} & \cellcolor{gray!10}{174} & \cellcolor{gray!10}{-1.6557612} & \cellcolor{gray!10}{0.8961531}\\
\addlinespace
PH- StableTumble, RL-StableTumble & 1.1480962 & 0.3356066 & 174 & 3.4209585 & 0.0069987\\
\cellcolor{gray!10}{NMC- StaticCWH, PH- StaticCWH} & \cellcolor{gray!10}{-2.7260146} & \cellcolor{gray!10}{0.3356066} & \cellcolor{gray!10}{174} & \cellcolor{gray!10}{-8.1226496} & \cellcolor{gray!10}{0.0000000}\\
NMC- StaticCWH vs RL-StaticCWH & 0.2920767 & 0.3356066 & 174 & 0.8702948 & 1.0000000\\
\cellcolor{gray!10}{PH-StaticCWH vs RL-StaticCWH} & \cellcolor{gray!10}{3.0180913} & \cellcolor{gray!10}{0.3356066} & \cellcolor{gray!10}{174} & \cellcolor{gray!10}{8.9929444} & \cellcolor{gray!10}{0.0000000}\\
\bottomrule
\toprule
Contrast & Estimate & SE & Df & t.ratio & p.value\\
\midrule
\cellcolor{gray!10}{NMC-StableTumble, NMC-StaticCWH} & \cellcolor{gray!10}{8.466667} & \cellcolor{gray!10}{9.84931} & \cellcolor{gray!10}{174} & \cellcolor{gray!10}{0.8596203} & \cellcolor{gray!10}{1.0000000}\\
PH-StableTumble, PH-StaticCWH & -22.700000 & 9.84931 & 174 & -2.3047299 & 1.0000000\\
\cellcolor{gray!10}{RL-StableTumble, RL-StaticCWH} & \cellcolor{gray!10}{39.766667} & \cellcolor{gray!10}{9.84931} & \cellcolor{gray!10}{174} & \cellcolor{gray!10}{4.0375077} & \cellcolor{gray!10}{0.0101190}\\
NMC-StableTumble, PH-StableTumble & -57.066667 & 9.84931 & 174 & -5.7939759 & 0.0000042\\
\cellcolor{gray!10}{NMC-StableTumble, RL-StableTumble} & \cellcolor{gray!10}{-19.600000} & \cellcolor{gray!10}{9.84931} & \cellcolor{gray!10}{174} & \cellcolor{gray!10}{-1.9899870} & \cellcolor{gray!10}{1.0000000}\\
\addlinespace
PH-StableTumble, RL-StableTumble & 37.466667 & 9.84931 & 174 & 3.8039888 & 0.0240333\\
\cellcolor{gray!10}{NMC-StaticCWH, PH-StaticCWH} & \cellcolor{gray!10}{-88.233333} & \cellcolor{gray!10}{9.84931} & \cellcolor{gray!10}{174} & \cellcolor{gray!10}{-8.9583260} & \cellcolor{gray!10}{0.0000000}\\
NMC-StaticCWH, RL-StaticCWH & 11.700000 & 9.84931 & 174 & 1.1879004 & 1.0000000\\
\cellcolor{gray!10}{PH-StaticCWH, RL-StaticCWH} & \cellcolor{gray!10}{99.933333} & \cellcolor{gray!10}{9.84931} & \cellcolor{gray!10}{174} & \cellcolor{gray!10}{10.1462265} & \cellcolor{gray!10}{0.0000000}\\
\bottomrule
\end{tabular}
\caption{\label{tab:unnamed-chunk-9}Post-hoc Pairwise Comparisons between Policy and Dynamic Type. The top table contains pairwise comparisons for ANOVA, the bottom table contains pairwise comparisons for ART.}
\end{table}

\subsubsection{Summary}\label{summary}

The results from both non-parametric and parametric methods are consistent and show a significant interaction effect between policy and RSO dynamic mode. In addition post-hoc pairwise comparisons show that fuel consumption for the RL policy is significantly lower in static CWH compared to stable tumble. However, no significant differences are observed between the other policies (PH and NMC) across the two RSO dynamic modes. Additionally, within both RSO dynamic modes, fuel consumption for PH is significantly higher than the other two policies(NMC and RL), while no significant differences are found between RL and NMC.

\subsection{Inspection Percentage}\label{sec: inspection-proportion}

\begin{figure}[!htb]
\centering
\includegraphics[width=.9\linewidth]{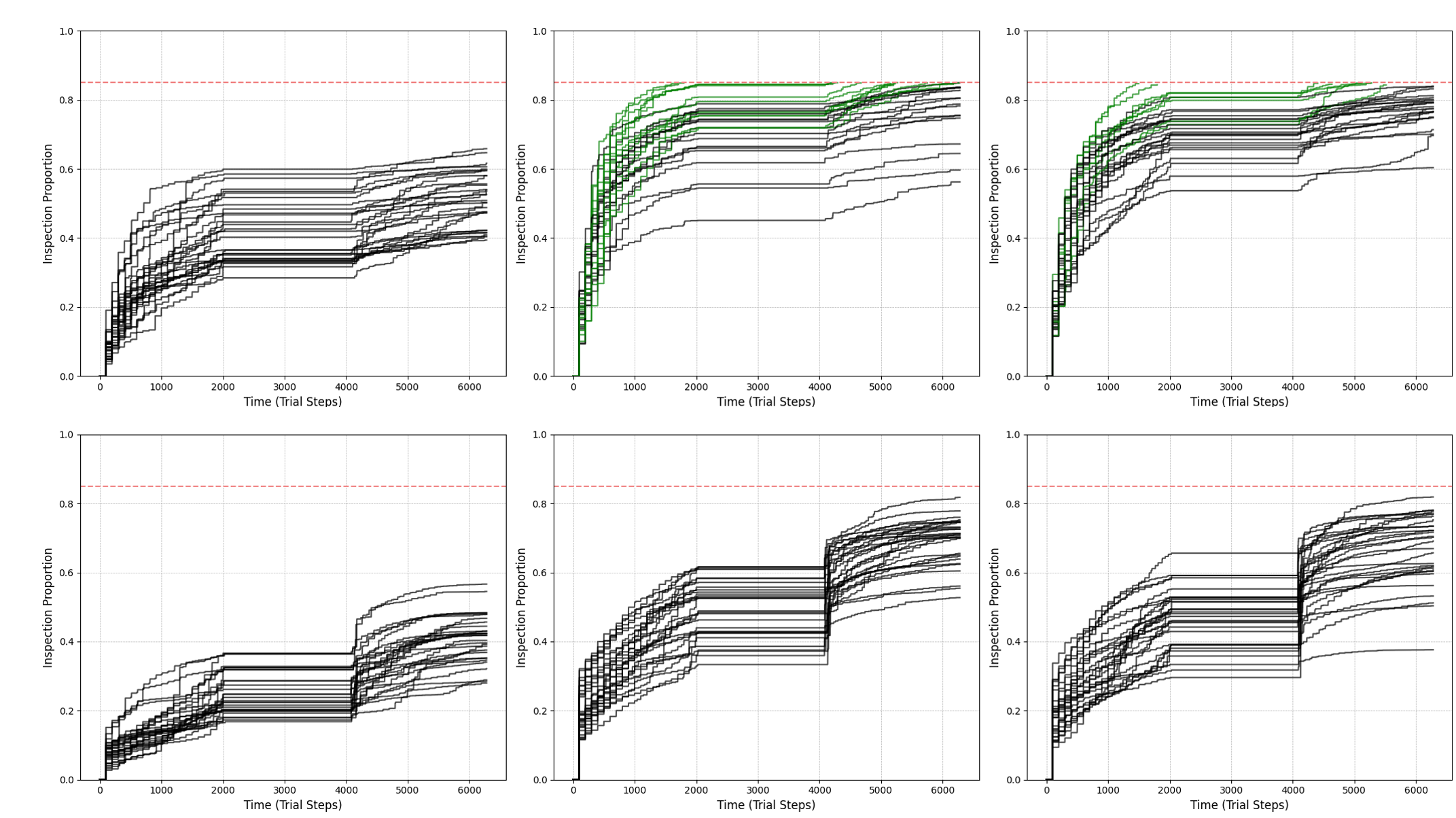}
\caption{Cumulative inspection progress over time for the three agent inspection mission across 30 trials. Specified by RSO dynamic mode and policy combination. Green traces indicate early mission success. The top panel is for the Stable Tumble mode while the bottom is for Static CWH.}
\label{fig: fuel_traj}
\end{figure}

\begin{figure}[h!]
\centering
\begin{subfigure}{0.7\linewidth}
\includegraphics[width=1\linewidth]{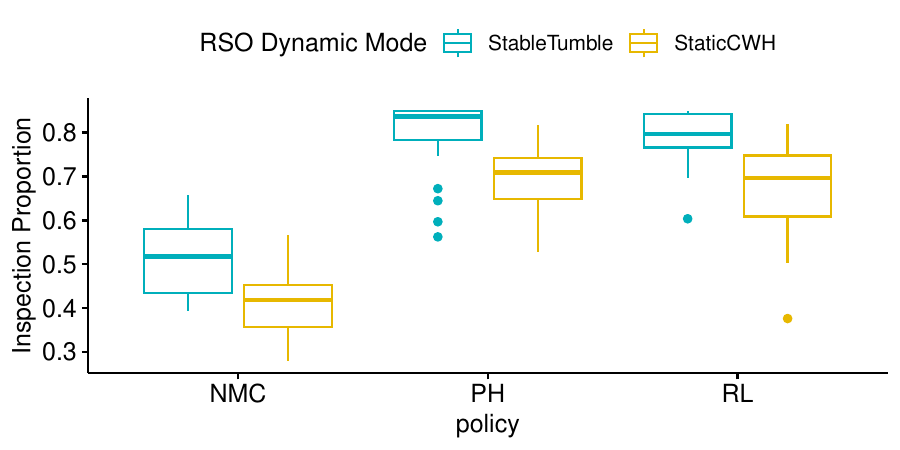}
\end{subfigure}
\begin{subfigure}{0.7\linewidth}
\includegraphics[width=1\linewidth]{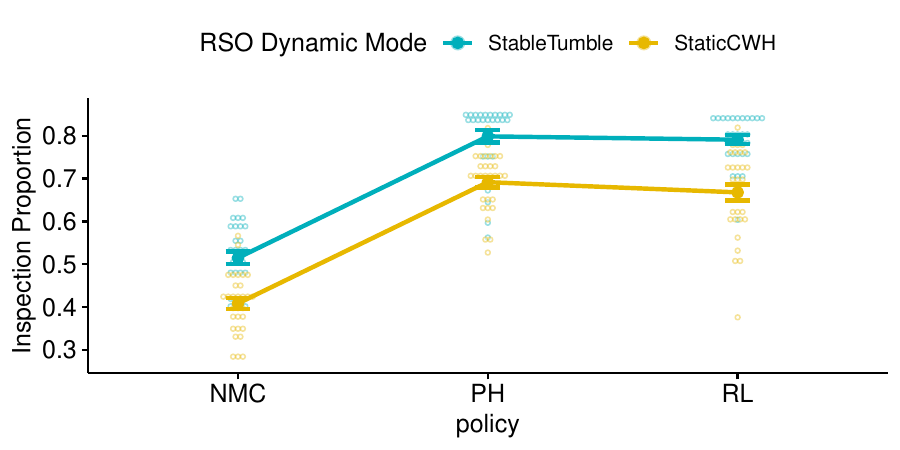}
\end{subfigure}
\caption{The upper panel shows box plot results for inspection proportion versus policy specification. The lower panel shows corresponding interaction effects.}
\label{fig: IPBP}
\end{figure}

There are two factors in the validation environment that greatly influence the difficulty of the inspection task: RSO geometry and Illumination. A baseline of $M=.85$ is set as a threshold for inspection completion; calculated by apriori estimation of the set of all POIs illuminated for static CWH and stable tumble dynamic modes. PH and NMC policies are generated under spherical RSO geometry, it is expected to see performance differential as a result of testing them under a new geometry (Aura). Heuristically, policies that guide deputies towards positions where the sun is consistently at their back will be rewarded in the information incentive; but may be penalized in the fuel incentive. The PH and NMC strategy places deputies in areas that will provide exposure to new information \textit{over time} - relying on predictive stability in the Monte Carlo simulation. Due to the rate at which sun position changes relative to the deputy, NMC and RL strategies may be difficult to plan retrieval opportunities where the action specification is increasingly sensitive to environmental changes. The stable tumble provides more opportunities for retrieval due to the rate of rotation and the variability in POIs that are illuminated. This is reflected by improved performance (in mean) of all policies, see Figure~\ref{fig: IPBP} and Table~\ref{tab:unnamed-chunk-12}. Both PH and RL outperformed NMC across all dynamic modes. Figure~\ref{fig: IPBP} does not exibit interaction effects between policy and RSO Dynamic Mode. This allows us to test significance related to each factor independently.

\begin{table}[!ht]
\centering
\begin{tabular}[t]{lllrrr}
\toprule
Dynamic & Policy & Variable & Sample Size & Mean & Standard Deviation\\
\midrule
\cellcolor{gray!10}{StableTumble} & \cellcolor{gray!10}{NMC} & \cellcolor{gray!10}{Inspection Proportion} & \cellcolor{gray!10}{30} & \cellcolor{gray!10}{0.515} & \cellcolor{gray!10}{0.079}\\
StaticCWH & NMC & Inspection Proportion & 30 & 0.409 & 0.071\\
\cellcolor{gray!10}{StableTumble} & \cellcolor{gray!10}{PH} & \cellcolor{gray!10}{Inspection Proportion} & \cellcolor{gray!10}{30} & \cellcolor{gray!10}{0.798} & \cellcolor{gray!10}{0.080}\\
StaticCWH & PH & Inspection Proportion & 30 & 0.692 & 0.069\\
\cellcolor{gray!10}{StableTumble} & \cellcolor{gray!10}{RL} & \cellcolor{gray!10}{Inspection Proportion} & \cellcolor{gray!10}{30} & \cellcolor{gray!10}{0.791} & \cellcolor{gray!10}{0.058}\\
\addlinespace
StaticCWH & RL & Inspection Proportion & 30 & 0.668 & 0.103\\
\bottomrule
\end{tabular}
\caption{\label{tab:unnamed-chunk-12}Summary statistics for inspection performance across
different combinations of RSO Dynamic Mode and policy.}
\end{table}

\subsubsection{Two-way ANOVA Test}\label{two-way-anova-test-1}

\begin{figure}
  \begin{minipage}[b]{.45\linewidth}
    \centering
    \begin{tabular}[t]{lrr}
    \toprule
    Variable & Statistic & p\\
    \midrule
    \cellcolor{gray!10}{residuals(model)} & \cellcolor{gray!10}{0.9584102} & \cellcolor{gray!10}{3.63e-05}\\
    \bottomrule
    \toprule
    (Df1, Df2) & Statistic & p\\
    \midrule
    \cellcolor{gray!10}{(5, 174)} & \cellcolor{gray!10}{2.145164} & \cellcolor{gray!10}{0.0623124}\\
    \bottomrule
    &&\\
    &&\\
    \end{tabular}
    \captionof{table}{Shapiro (top) and Levene's (bottom) test results conducted on fuel utilization.}\label{tab:ANOVA_assump_insp}
  \end{minipage}
  \hfill
  \begin{minipage}[b]{.45\linewidth}
    \centering
    \includegraphics[width=0.7\linewidth]{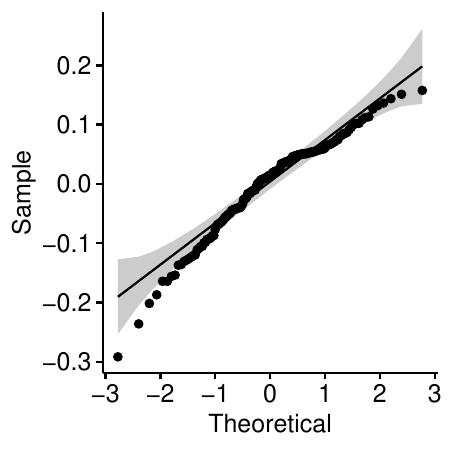}
    \captionof{figure}{Quantile-Quantile (QQ) plot for sampled trials against inspection percentage.}\label{fig: QQplotIP}
  \end{minipage}
\end{figure}

A Two-Way ANOVA test was conducted to determine if there is a significant impact of RSO Dynamic Mode and Policy on inspection performance while also characterizing potential interaction effects. Before beginning, we test the relevant assumptions of normality and homogeneity of variance. Both the significant p-value in a Shapiro test (Table~\ref{tab:ANOVA_assump_insp}) and the QQ plot show that the assumption of normality is not met; see Figure~\ref{fig: QQplotFuel}. Levene's test shows that there is not enough evidence to conclude that there is a significant difference in variance across the groups. As such, the homogeneity of variances assumption has been met; see Table~\ref{tab:ANOVA_assump_insp}. The analysis of variance on inspection proportion showed statistically significant main effects of both policy and RSO dynamic mode, but no significant interaction between them. This indicates that both policy and RSO dynamic mode have significant effects on inspection proportion; see Table~\ref{tab:unnamed-chunk-13}.

\begin{table}[!ht]
\centering
\begin{tabular}[t]{lrrrrr}
\toprule
  & Df & Sum Sq & Mean Sq & F value & Pr(>F)\\
\midrule
\cellcolor{gray!10}{policy} & \cellcolor{gray!10}{2} & \cellcolor{gray!10}{3.0400735} & \cellcolor{gray!10}{1.5200368} & \cellcolor{gray!10}{251.2414759} & \cellcolor{gray!10}{0.0000000}\\
dynamic & 1 & 0.5648779 & 0.5648779 & 93.3666662 & 0.0000000\\
\cellcolor{gray!10}{policy:dynamic} & \cellcolor{gray!10}{2} & \cellcolor{gray!10}{0.0026681} & \cellcolor{gray!10}{0.0013340} & \cellcolor{gray!10}{0.2204995} & \cellcolor{gray!10}{0.8023418}\\
Residuals & 174 & 1.0527179 & 0.0060501 & NA & NA\\
\bottomrule
\end{tabular}
\caption{\label{tab:unnamed-chunk-13}ANOVA Summary for inspection proportion.}
\end{table}

\subsubsection{Aligned Rank Transform and Pairwise Comparisons}\label{non-parametric-alternative-aligned-rank-transform-art-1}

The ART method results shows statistically significant main effects of
both policy and RSO dynamic mode on inspection proportion, with no
significant interaction between them. This suggests that both policy and
RSO dynamic mode impact inspection proportion. Post-hoc pairwise
comparisons with Bonferroni adjustment were conducted to identify
specific differences between policy types. The results show significant differences in adjusted means between NMC and RL, as well as between NMC and PH. However, there is no significant difference between the PH and RL policies; see Table~\ref{tab:unnamed-chunk-18} for more information.

\begin{table}[!ht]
\centering
\begin{tabular}[t]{lrrrrrr}
\toprule
  & Df & Df.res & Sum Sq & Sum Sq.res & F value & Pr(>F)\\
\midrule
\cellcolor{gray!10}{policy} & \cellcolor{gray!10}{2} & \cellcolor{gray!10}{174} & \cellcolor{gray!10}{3.104425e+05} & \cellcolor{gray!10}{174444.1} & \cellcolor{gray!10}{154.8261338} & \cellcolor{gray!10}{0.0000000}\\
dynamic & 1 & 174 & 1.865024e+05 & 297926.8 & 108.9241433 & 0.0000000\\
\cellcolor{gray!10}{policy:dynamic} & \cellcolor{gray!10}{2} & \cellcolor{gray!10}{174} & \cellcolor{gray!10}{6.711111e+00} & \cellcolor{gray!10}{485247.5} & \cellcolor{gray!10}{0.0012032} & \cellcolor{gray!10}{0.9987975}\\
\bottomrule
\end{tabular}
\caption{\label{tab:unnamed-chunk-17}ANOVA results based on Aligned Rank Transform for inspection proportion.}
\centering
\begin{tabular}[t]{lrrrrr}
\toprule
Contrast & Estimate & SE & df & t.ratio & p.value\\
\midrule
\cellcolor{gray!10}{NMC - PH} & \cellcolor{gray!10}{-0.2832387} & \cellcolor{gray!10}{0.0142011} & \cellcolor{gray!10}{174} & \cellcolor{gray!10}{-19.944902} & \cellcolor{gray!10}{0.0000000}\\
NMC - RL & -0.2674514 & 0.0142011 & 174 & -18.833200 & 0.0000000\\
\cellcolor{gray!10}{PH - RL} & \cellcolor{gray!10}{0.0157874} & \cellcolor{gray!10}{0.0142011} & \cellcolor{gray!10}{174} & \cellcolor{gray!10}{1.111703} & \cellcolor{gray!10}{0.8033993}\\
\bottomrule
\toprule
Contrast & Estimate & SE & Df & t.ratio & p.value\\
\midrule
\cellcolor{gray!10}{NMC - PH} & \cellcolor{gray!10}{-92.76667} & \cellcolor{gray!10}{5.780865} & \cellcolor{gray!10}{174} & \cellcolor{gray!10}{-16.047194} & \cellcolor{gray!10}{0.0000000}\\
NMC - RL & -82.53333 & 5.780865 & 174 & -14.276986 & 0.0000000\\
\cellcolor{gray!10}{PH - RL} & \cellcolor{gray!10}{10.23333} & \cellcolor{gray!10}{5.780865} & \cellcolor{gray!10}{174} & \cellcolor{gray!10}{1.770208} & \cellcolor{gray!10}{0.5480243}\\
\bottomrule
\end{tabular}
\caption{\label{tab:unnamed-chunk-18} Posthoc Pairwise Comparisons Between Policy Types. The top table contains pairwise comparisons for ANOVA, the bottom table contains pairwise comparisons for ART. As compared to Table~\ref{tab:unnamed-chunk-9}, there are fewer comparisons due to the lack of a significant interaction effect.}
\end{table}

\subsubsection{Summary}\label{summary-1}

The results from both non-parametric and parametric methods are consistent, showing a significant effect of both policy and RSO dynamic mode on inspection proportion. On average, the StableTumble RSO dynamic mode has a higher inspection proportion compared to StaticCWH. Additionally, post-hoc pairwise comparisons for policy levels indicate that NMC has a lower inspection proportion compared to RL and PH, while there is no significant difference between RL and PH.

\section{Conclusion}
This work investigated the relationship between policy type and RSO dynamic mode on inspection completion. Conducted in a purpose-built validation environment, results are robust to model overfitting and help assess stability against primitive modeling assumptions. The results from Section~\ref{sec: fuel-consumption} and Section~\ref{sec: inspection-proportion} taken together indicate a significant performance increase across both RSO dynamic modes and policy specification in favor of the active/RL policy. The inspection performance of NMC was significantly degraded under the addition of more complex illumination and RSO surface geometry. Although significantly more fuel efficient than the PH, the NMC specification still required comparable fuel usage to the RL policy when entrance and trajectory maintenance criteria were enforced. The PH strategy did exceptionally well in inspection performance with deputies strategically place to maximize alignment over time with the source of illumination. This comes at the cost of high fuel usage for position maintenance. The active policy specification allows for unanticipated conditions to be corrected in real-time. This makes them advantageous for scenarios or environments that are uncertain. In contrast, the passively defined strategies are predictable and simple allowing for enhanced safety and control over the spacecraft. 
\cite{Lei25}
\section*{ACKNOWLEDGMENT}
S. Phillips was funded directly by the Air Force Research Laboratory (AFRL) as Research Staff. J. Aurand, C. Pang, S. Mokhtar, H. Lei, and S. Cutlip were funded by AFRL under FA9453-21-C-0602.

\bibliography{Biblio}

\begin{thebibliography}{33}
\newcommand{\enquote}[1]{``#1''}
\providecommand{\natexlab}[1]{#1}
\providecommand{\url}[1]{\texttt{#1}}
\providecommand{\urlprefix}{URL }
\expandafter\ifx\csname urlstyle\endcsname\relax
  \providecommand{\doi}[1]{\discretionary{}{}{}https://doi.org/#1}\else
  \providecommand{\doi}[1]{\discretionary{}{}{}\urlstyle{rm}\url{https://doi.org/#1}}\fi

\bibitem[{Lei et~al.(2022)Lei, Shubert, Damron, Lang, and Phillips}]{Lei22}
Lei, H.~H., Shubert, M., Damron, N., Lang, K., and Phillips, S., \enquote{Deep
  Reinforcement Learning for Multi-Agent Autonomous Satellite Inspection,}
  \emph{44th AAS Guidance, Navigation and Control Conference}, American
  Astronautical Society Rocky Mountain Region, 2022.

\bibitem[{Bernhard et~al.(2020{\natexlab{a}})Bernhard, Choi, Rahmani, Chung,
  and Hadaegh}]{Bernhard20}
Bernhard, B., Choi, C., Rahmani, A., Chung, S.-J., and Hadaegh, F.,
  \enquote{Coordinated Motion Planning for On-Orbit Satellite Inspection using
  a Swarm of Small-Spacecraft,} \emph{2020 IEEE Aerospace Conference},
  2020{\natexlab{a}}, pp. 1--13.
\newblock \doi{10.1109/AERO47225.2020.9172747}.

\bibitem[{Nakka et~al.(2021{\natexlab{a}})Nakka, Hönig, Choi, Harvard,
  Rahmani, and Chung}]{Nakka21}
Nakka, Y. K.~K., Hönig, W., Choi, C., Harvard, A., Rahmani, A., and Chung,
  S.-J., \enquote{Information-Based Guidance and Control Architecture for
  Multi-Spacecraft On-Orbit Inspection,} \emph{AIAA Scitech Forum},
  2021{\natexlab{a}}.
\newblock \doi{10.2514/6.2021-1103},
  \urlprefix\url{https://arc.aiaa.org/doi/abs/10.2514/6.2021-1103}.

\bibitem[{Phillips et~al.(2022)Phillips, Petersen, and Fierro}]{Phillips2022}
Phillips, S., Petersen, C., and Fierro, R., \enquote{Robust, Resilient, and
  Energy-Efficient Satellite Formation Control,} \emph{Intelligent Control and
  Smart Energy Management: Renewable Resources and Transportation}, edited by
  M.~J. Blondin, J.~P. Fernandes~Trov{\~a}o, H.~Chaoui, and P.~M. Pardalos,
  Springer International Publishing, Cham, 2022, pp. 223--251.

\bibitem[{Aurand et~al.(2024)Aurand, Cutlip, Lei, Lang, and Phillips}]{JSR24}
Aurand, J., Cutlip, S., Lei, H., Lang, K., and Phillips, S., \enquote{Deep
  Q-Learning for Decentralized Multi-Agent Inspection of a Tumbling Target,}
  \emph{Journal of Spacecraft and Rockets}, Vol.~61, No.~2, 2024, pp. 341--354.
\newblock \doi{10.2514/1.A35749},
  \urlprefix\url{https://doi.org/10.2514/1.A35749}.

\bibitem[{Oestreich et~al.(2021)Oestreich, Espinoza, Todd, Albee, and
  Linares}]{oestreich21}
Oestreich, C., Espinoza, A.~T., Todd, J., Albee, K., and Linares, R.,
  \enquote{On-Orbit Inspection of an Unknown, Tumbling Target Using NASA's
  Astrobee Robotic Free-Flyers,} \emph{Proceedings of the IEEE/CVF Conference
  on Computer Vision and Pattern Recognition}, 2021, pp. 2039--2047.

\bibitem[{Dor and Tsiotras(2018)}]{Dor18}
Dor, M., and Tsiotras, P., \enquote{ORB-SLAM Applied to Spacecraft
  Non-Cooperative Rendezvous,} \emph{2018 Space Flight Mechanics Meeting},
  2018.
\newblock \doi{10.2514/6.2018-1963},
  \urlprefix\url{https://arc.aiaa.org/doi/abs/10.2514/6.2018-1963}.

\bibitem[{Hays et~al.(2023)Hays, Miller, Soderlund, Phillips, and
  Henderson}]{hays2023}
Hays, C., Miller, K., Soderlund, A., Phillips, S., and Henderson, T.,
  \enquote{Autonomous Local Catalog Maintenance of Close Proximity Satellite
  Systems on Closed Natural Motion Trajectories,} 2023.
\newblock \doi{10.48550/arXiv.2302.00601}.

\bibitem[{Grange et~al.(2023)Grange, Sandu, Soderlund, and
  Phillips}]{GrangePLANS2023}
Grange, D., Sandu, R., Soderlund, A.~A., and Phillips, S., \enquote{Consensus
  over Region-Based Posed Estimation for Satellites,} \emph{IEEE/ION Position
  Location and Navigation Symposium}, 2023.

\bibitem[{Woffinden and Geller(2007)}]{woffinden07}
Woffinden, D.~C., and Geller, D.~K., \enquote{Relative angles-only navigation
  and pose estimation for autonomous orbital rendezvous,} \emph{Journal of
  Guidance, Control, and Dynamics}, Vol.~30, No.~5, 2007, pp. 1455--1469.

\bibitem[{Maestrini and Di~Lizia(2022)}]{Maestrini22}
Maestrini, M., and Di~Lizia, P., \enquote{Guidance Strategy for Autonomous
  Inspection of Unknown Non-Cooperative Resident Space Objects,} \emph{Journal
  of Guidance, Control, and Dynamics}, Vol.~45, No.~6, 2022, pp. 1126--1136.
\newblock \doi{10.2514/1.G006126},
  \urlprefix\url{https://doi.org/10.2514/1.G006126}.

\bibitem[{Miller et~al.()Miller, Phillips, and Soderlund}]{miller2023SciTech}
Miller, K., Phillips, S., and Soderlund, A.~A., \enquote{Multi-agent Control of
  Chaser Satellites using Games with Lexicographic Preferences,} \emph{AIAA
  SCITECH 2023 Forum}, ????
\newblock \doi{10.2514/6.2023-2675}.

\bibitem[{Vries and Phillion(2010)}]{Vries2010}
Vries, W. H.~D., and Phillion, D.~W., \enquote{Monte Carlo Method for Collision
  Probability Calculations Using 3D Satellite Models,} \emph{Advanced Maui
  Optical and Space Survelliance Technologies Conference}, 2010.

\bibitem[{Omar et~al.(2021)Omar, Riano-Rios, and Bevilacqua}]{Omar2021}
Omar, S.~R., Riano-Rios, C., and Bevilacqua, R., \enquote{The Drag Maneuvering
  Device for the Semi-Passive Three-Axis Attitude Stabilization of Low Earth
  Orbit Nanosatellites,} \emph{Journal of Small Satellites}, Vol.~10, 2021, p.
  943.

\bibitem[{Stesina(2021)}]{Stesina2021}
Stesina, F., \enquote{Tracking model predictive control for docking maneuvers
  of a CubeSat with a big spacecraft,} \emph{Aerospace}, Vol.~8, 2021.
\newblock \doi{10.3390/aerospace8080197}.

\bibitem[{Albee et~al.(2021)Albee, Oestreich, Specht, Espinoza, Todd, Hokaj,
  Lampariello, and Linares}]{Albee2021}
Albee, K., Oestreich, C., Specht, C., Espinoza, A.~T., Todd, J., Hokaj, I.,
  Lampariello, R., and Linares, R., \enquote{A Robust Observation, Planning,
  and Control Pipeline for Autonomous Rendezvous with Tumbling Targets,}
  \emph{Frontiers in Robotics and AI}, Vol.~8, 2021.
\newblock \doi{10.3389/frobt.2021.641338}.

\bibitem[{Zampato et~al.(2013)Zampato, Pistellato, and Maddalena}]{Zampato2013}
Zampato, M., Pistellato, R., and Maddalena, D., \enquote{Visual Motion
  Estimation for Tumbling Satellite Capture,} \emph{British Machine Vision
  Conference}, 2013.
\newblock \doi{10.5244/c.10.68}.

\bibitem[{Lampariello(2021)}]{Lampariello2021}
Lampariello, R., \enquote{Optimal Motion Planning for Object Interception and
  Capture,} \emph{Ph.D Thesis at Technical University of Darmstadt}, 2021.

\bibitem[{Setterfield(2017)}]{Setterfield2017}
Setterfield, T.~P., \enquote{On-Orbit Inspection of a Rotating Object Using a
  Moving Observer,} \emph{Ph.D Thesis in the Department of Aeronautics and
  Astronautics at Massachusetts Institute of Technology}, 2017.

\bibitem[{Nakka et~al.(2021{\natexlab{b}})Nakka, Hönig, Choi, Harvard,
  Rahmani, and Chung}]{Nakka2021}
Nakka, Y.~K., Hönig, W., Choi, C., Harvard, A., Rahmani, A., and Chung, S.~J.,
  \enquote{Information-based guidance and control architecture for
  multi-spacecraft on-orbit inspection,} \emph{AIAA Scitech 2021 Forum},
  2021{\natexlab{b}}, pp. 1--22.
\newblock \doi{10.2514/6.2021-1103}.

\bibitem[{Bernhard et~al.(2020{\natexlab{b}})Bernhard, Choi, Rahmani, Chung,
  and Hadaegh}]{Bernhard2020}
Bernhard, B., Choi, C., Rahmani, A., Chung, S., and Hadaegh, F.,
  \enquote{Coordinated Motion Planning for On-Orbit Satellite Inspection using
  a Swarm of Small-Spacecraft,} \emph{IEEE Aerospace Conference},
  2020{\natexlab{b}}.

\bibitem[{Elipe et~al.(1997)Elipe, Arribas, and Riaguas}]{Elipe1997}
Elipe, A., Arribas, M., and Riaguas, A., \enquote{Complete analysis of
  bifurcations in the axial gyrostat problem,} \emph{Journal of Physics A:
  Mathematical and General}, Vol.~30, 1997, pp. 587--601.
\newblock \doi{10.1088/0305-4470/30/2/021}.

\bibitem[{Richter(2006)}]{Richter2006}
Richter, P.~H., \enquote{Regular and Chaotic Rigid Body Dynamics,}
  \emph{Nonlinear Phenomena in Complex Systems}, 2006.

\bibitem[{Cutlip et~al.(2024)Cutlip, Aurand, Lang, and Phillips}]{SC2024}
Cutlip, S., Aurand, J., Lang, K., and Phillips, S., \enquote{Multi-Agent
  Passive Inspection Coverage of an Unknown Torque-Free Rigid Body Using Monte
  Carlo Analysis and Quaternion Measurements,} \emph{AIAA SciTech}, 2024.
\newblock \urlprefix\url{https://arc.aiaa.org/doi/epdf/10.2514/6.2024-2882}.

\bibitem[{Phillips et~al.()Phillips, Lippay, Baker, Soderlund, and
  Shubert}]{Phillips_ST24}
Phillips, S., Lippay, Z., Baker, D., Soderlund, A.~A., and Shubert, M.,
  \emph{Emulation of Close-Proximity Spacecraft Dynamics in Terrestrial
  Environments Using Unmanned Aerial Vehicles}, ????
\newblock \doi{10.2514/6.2024-1207},
  \urlprefix\url{https://arc.aiaa.org/doi/abs/10.2514/6.2024-1207}.

\bibitem[{van Wijk et~al.(2023)van Wijk, Dunlap, and Hobbs}]{vanWijk23}
van Wijk, D., Dunlap, K., and Hobbs, K., \enquote{Deep Reinforcement Learning
  for Autonomous Spacecraft Inspection using Illumination,} \emph{AAS/AIAA
  Astrodynamics Specialist Conference, Big Sky, Montana}, 2023.

\bibitem[{Katz et~al.(2007)Katz, Tal, and Basri}]{Katz07}
Katz, S., Tal, A., and Basri, R., \enquote{Direct visibility of point sets,}
  \emph{ACM SIGGRAPH papers}, 2007, pp. 24--es.

\bibitem[{Irvin(2007)}]{Irvin2007}
Irvin, D.~J., \enquote{Optimal Control Strategies for Constrained Relative
  Orbits,} Ph.D. thesis, Air Force Institute of Technology, 2007.

\bibitem[{Hughes(2004)}]{Peter04}
Hughes, P., \emph{Spacecraft Attitude Dynamics}, Courier Corporation, 2004.

\bibitem[{Markley and Crassidis(2014)}]{FMark}
Markley, F.~L., and Crassidis, J.~L., \enquote{Space Technology Library
  Fundamentals of Spacecraft Attitude Determination and Control,}
  \emph{Springer}, 2014.

\bibitem[{Mnih et~al.(2013)Mnih, Kavukcuoglu, Silver, Graves, Antonoglou,
  Wierstra, and Riedmiller}]{Mnih13}
Mnih, V., Kavukcuoglu, K., Silver, D., Graves, A., Antonoglou, I., Wierstra,
  D., and Riedmiller, M., \enquote{Playing atari with deep reinforcement
  learning,} \emph{NIPS Deep Learning Workshop}, 2013.
\newblock \doi{10.48550/arXiv.1312.5602}.

\bibitem[{Mnih et~al.(2015)Mnih, Kavukcuoglu, Silver, Rusu, Veness, Bellemare,
  Graves, Riedmiller, Fidjeland, Ostrovski et~al.}]{Mnih15}
Mnih, V., Kavukcuoglu, K., Silver, D., Rusu, A.~A., Veness, J., Bellemare,
  M.~G., Graves, A., Riedmiller, M., Fidjeland, A.~K., Ostrovski, G., et~al.,
  \enquote{Human-level control through deep reinforcement learning,}
  \emph{Nature}, Vol. 518, No. 7540, 2015, pp. 529--533.

\bibitem[{Lei et~al.(2025)Lei, Lippay, Zaman, Aurand, Maghareh, and
  Phillips}]{Lei25}
Lei, H., Lippay, Z., Zaman, A., Aurand, J., Maghareh, A., and Phillips, S.,
  \enquote{Stability Analysis of Deep Reinforcement Learning for Multi-Agent
  Inspection in a Terrestrial Testbed,} \emph{AIAA Scitech Forum}, 2025.

\end{thebibliography}

\end{document}